%% file: main.tex
\documentclass[10pt,twocolumn,letterpaper]{article}

\usepackage[pagenumbers]{cvpr} 

\input{preamble}

\definecolor{cvprblue}{rgb}{0.21,0.49,0.74}
\usepackage[pagebackref,breaklinks,colorlinks,allcolors=cvprblue]{hyperref}
\usepackage{multirow}
\usepackage{pgfplots}
\usepackage{times}
\usepackage{pgfplotstable}
\pgfplotsset{compat=1.18}
\usepackage{xcolor}
\usepackage{colortbl}
\usepackage{adjustbox}
\usepackage{makecell}
\usepackage{algorithm}
\usepackage{algpseudocode}
\usepackage{pifont}
\usepackage[breakable]{tcolorbox}
\usepackage{dblfloatfix} 
\usepackage{enumitem}
\usepackage{placeins}

\colorlet{PastaYellow}{yellow!10}
\definecolor{darkgreen}{HTML}{006400}
\definecolor{deltared}{HTML}{C0392B}
\newcommand{\up}[1]{$\scriptstyle\textcolor{darkgreen}{(\textbf{+#1})}$}
\newcommand{\std}[1]{$_{\pm #1}$}
\newcommand{\berv}{\mathbb{V}}

\makeatletter
\renewcommand\paragraph{\@startsection{paragraph}{4}{\z@}{0.8ex \@plus0.1pt \@minus.5ex}{-0.2em}{\normalfont\normalsize\bfseries}}
\makeatother

\newcommand{\ours}{Search-RSI\xspace}
\newcommand{\searchworld}{\mathcal{W}}
\newcommand{\taskpool}{\mathcal{D}}
\newcommand{\knowledge}{\mathcal{K}}
\newcommand{\frontier}{\mathcal{F}}
\newcommand{\candset}{\mathcal{C}}
\newcommand{\admitset}{\mathcal{S}}
\newcommand{\demoset}{\mathcal{B}}
\newcommand{\unitspace}{\mathcal{U}}
\newcommand{\score}{\mathcal{Q}}
\newcommand{\Cons}{\mathrm{Cons}}
\newcommand{\Kstg}{\knowledge^{\mathrm{stg}}}
\newcommand{\Keval}{\knowledge^{\mathrm{eval}}}
\newcommand{\hyp}[2]{#2}

\title{{\fontsize{13.2pt}{16pt}\selectfont Search-RSI: Expanding Search Worlds Along Multimodal Agents' Capability Frontier}}
\author{%
Chengchao Lv$^{1}$ \quad Yin Wang$^{2}$ \quad Chunzheng Zhu$^{1}$ \quad
Wanli Chang$^{1}$\thanks{Corresponding author: \texttt{changwanli@hnu.edu.cn}.}\\[3pt]
$^{1}$Hunan University \quad $^{2}$New York University
}

\usepackage[final]{microtype}
\AtBeginDocument{\setlength{\parfillskip}{0pt plus 0.55\columnwidth}}
\allowdisplaybreaks

\hypersetup{
pdftitle={Search-RSI: Expanding Search Worlds Along Multimodal Agents' Capability Frontier},
pdfauthor={Chengchao Lv, Yin Wang, Chunzheng Zhu, Wanli Chang}
}

\begin{document}
\maketitle

\begin{abstract}
Multimodal search agents answer knowledge intensive visual questions by grounding them in retrieved evidence, and recent recipes train them on multi hop tasks that are synthesized from knowledge graphs and grown online to keep pace with the agent. These recipes, however, treat the searchable world as fixed: a new question can be well formed and verifiable while its evidence is absent from the retrieval backend, so the agent issues the right query, retrieves nothing, and is penalized as if its policy were at fault. Such unreachable negatives accumulate with every round of growth. We propose \textbf{\ours}, a recursive self improvement framework that expands the search world along the agent's capability frontier. In each round, the current policy diagnoses its partially mastered regions of a task space spanned by hop depth, relation rarity, and distractor density; new tasks there are grown together with their evidence; and the same policy decides how they are used: candidates on which its probes disagree feed reinforcement learning, while those beyond its reach become hint guided demonstrations of its own. Keeping the policy inside this loop matters: an improver frozen at the start of training loses \hyp{abs.frozen}{3.2} points. On nine multimodal search benchmarks, \ours keeps improving where static task growth saturates, raising the average accuracy of the recipe it builds on by \hyp{abs.d9}{9.8} and \hyp{abs.d27}{9.5} points at 9B and 27B, of which \hyp{abs.directed}{7.0} points are out of reach for a world that is merely made consistent. The loop uses no external model and only privileged task structure that is removed before training, at $1.27\times$ the compute of static growth.
\end{abstract}

\section{Introduction}
\begin{figure}[!t]
    \centering
    \IfFileExists{images/intro-motivation.pdf}{%
        \includegraphics[width=\linewidth]{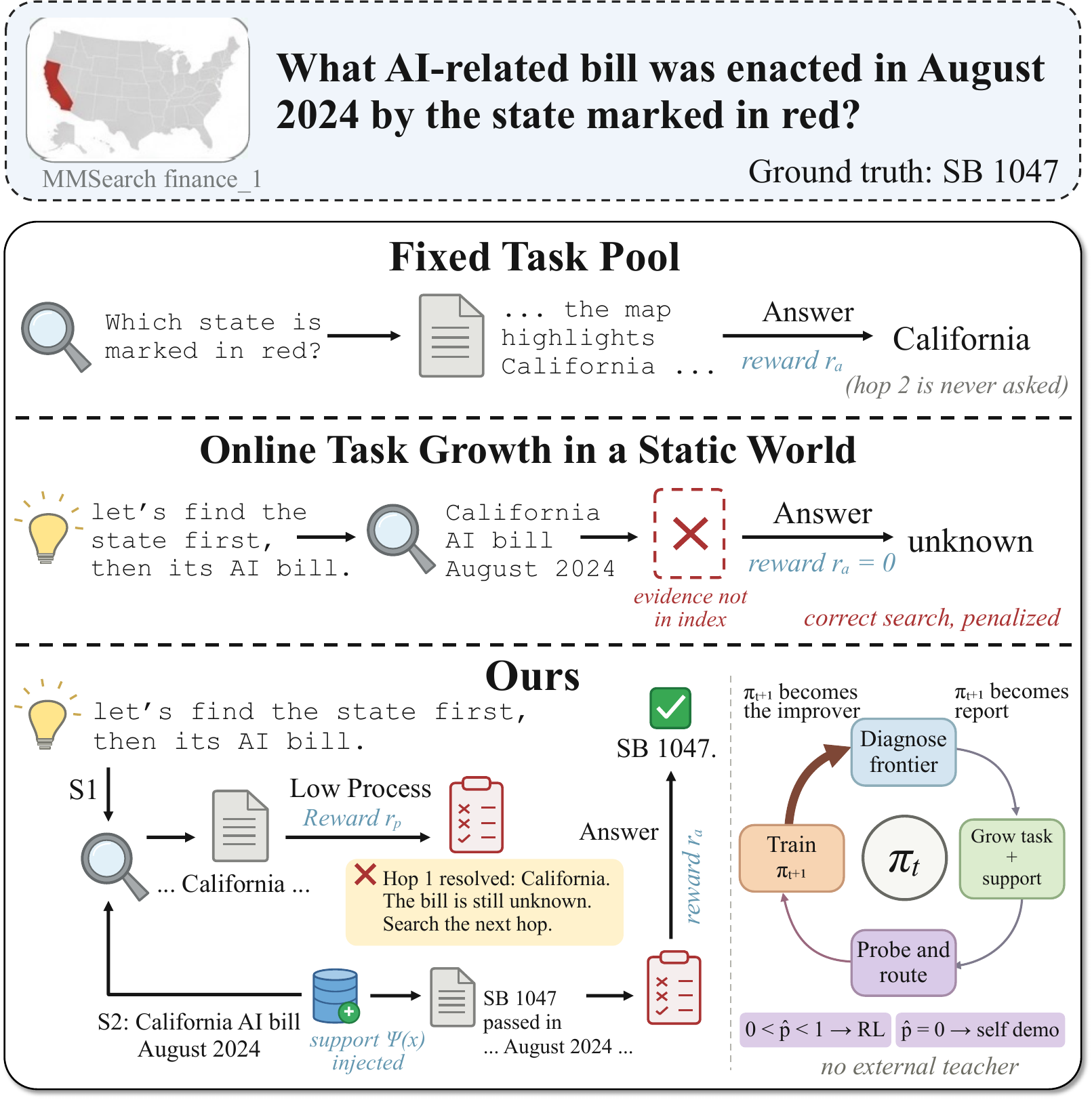}%
    }{%
        \fbox{\parbox[c][3.5in][c]{0.94\linewidth}{\centering
        \textbf{Question box} (MMSearch finance\_1)\\[4pt]
        Fixed Task Pool\\[2pt] - - - - - - - - - -\\[2pt]
        Online Task Growth in a Static World\\[2pt] - - - - - - - - - -\\[2pt]
        Ours: trajectory (left) + recursive loop (right)\\[6pt]
        (icon legend)}}%
    }
    \caption{\textbf{Fixed pool and static world growth (top) vs.\ \ours (bottom)} on an \textsc{MMSearch} question. A fixed pool never asks the second hop; static growth asks it without its supporting fact, so a correct search is penalized. \ours injects the support of the task, and the improved policy $\pi_{t+1}$ diagnoses, admits, and demonstrates the tasks of the next round.}
    \label{fig:motivation}
    \vspace{-3mm}
\end{figure}

One of the central drivers behind the rapid progress of multimodal large language models (MLLMs) has been their ability to \textit{search}: to ground an answer in external evidence by recognizing visual entities, issuing queries, and aggregating what is retrieved \cite{jiang2024mmsearch,cheng2025simplevqa,huang2026vision,zeng2026vision,fu2026seeking}. Consider the \textsc{MMSearch} question in Fig.~\ref{fig:motivation}, which shows a map of the United States with one state marked in red and asks which AI related bill that state enacted in August 2024: the agent must first recover the state and then search for a fact that no pretrained model has memorized. A commonly used approach is to train a multimodal search agent with supervised warm up and RL on retrieval actions, supplied with multi hop questions synthesized from a knowledge graph \cite{jin2025search,wu2025mmsearch,geng2026webwatcher,li2026onesearch,tang2026long}, and open recipes now sample graph paths at scale and render them into image grounded questions that continuously feed training \cite{chen2026opensearch,jiao2026searcheyes,huang2026towards}.

The standard paradigm for developing search agents has thus shifted from training on a fixed pool to \textit{online task growth}, where new tasks are generated as the agent improves so that it always faces questions it has not yet mastered. Building on this progress, a more sophisticated line of research couples growth with curricula and self evolution that propose tasks at the edge of competence \cite{portelas2020teacher,zhao2026absolute,dong2026agent,lu2026search}. However, this line of progress faces a critical bottleneck: the world that the agent searches in, namely the retrieval backend that decides which evidence is retrievable, is still treated as fixed. A synthesized question can be semantically valid and carry a verifiable answer while a fact on its supporting path is absent from the active index. The agent then issues the right query, retrieves nothing, and is penalized as if its policy were at fault (see Fig.~\ref{fig:motivation}). These \textit{unreachable negatives} are not minor: in our pilot, \hyp{intro.unsup}{29.0}\% of newly grown tasks lack a supporting element, and such samples come to account for \hyp{intro.un}{41.0}\% of all negative rewards, undermining the promise of online growth.

Existing solutions do not remove this bottleneck. Loading the entire knowledge graph into the index floods retrieval with distractors and shifts the difficulty of every existing task, while filtering prompts on which all rollouts fail \cite{yu2026dapo} or verifying tasks by execution and voting \cite{zhao2026absolute,huang2026r} either discards hard yet repairable tasks or assumes a world that is complete by construction. The difficulty is compounded by two further gaps. Growth schedules decide how many tasks to add but rarely where, and a scalar success rate treats tasks that fail for unrelated reasons, such as a late hop that breaks, a malformed query on a rare relation, or a look alike entity, as interchangeable, so new tasks land where the agent is already saturated or still hopeless. Moreover, even when the evidence exists, RL alone cannot learn from a task on which every rollout fails, because all advantages in the group vanish; the long and rare tasks that growth produces are exactly the ones it cannot use. We are thus left with a crucial question: \textit{how can we keep growing tasks at the agent's frontier while every new task stays answerable and learnable?}

In this work, we challenge the underlying hypothesis that the search world is a fixed backdrop of training. We propose a different perspective: the world itself is a training variable that should grow in step with the agent, in a recursive loop where the current policy decides what the world adds next and how the next policy learns from it. This insight forms the basis of our approach: \textbf{\ours}, a recursive self improvement approach that expands the search world along the agent's capability frontier. Each round diagnoses where the policy is partially competent, grows tasks there together with the evidence they need, and lets short probe rollouts of the policy itself route every candidate, so the improvement of one round reaches the mechanism that produces the next. The loop uses no external model: the only extra information is the path of the task itself, given as a hint during retries and removed before training, so it exists wherever tasks are synthesized and never at test time.

We validate \ours at two model scales on nine benchmarks, building the training world on the open \textsc{SearchEyes} knowledge graph \cite{jiao2026searcheyes} so that round 0 reproduces a published recipe. Expanding the world with the tasks keeps the agent improving where static growth saturates (\textit{e.g.,} \hyp{intro.s9}{59.3}$\rightarrow$\hyp{intro.o9}{69.1} at 9B), and the improver itself grows stronger: the share of out of reach tasks it can demonstrate rises from \hyp{intro.y0}{17}\% to \hyp{intro.y1}{41}\%, where a frozen copy of the initial policy stays flat. 
Our contributions are as follows. \textit{i)} We identify \textit{unreachable negatives} as a systematic failure of online task growth: a task can be verifiable while its evidence is absent from the retrieval backend, so a correct search is penalized and the training signal decays as the pool grows. \textit{ii)} We propose \ours, which treats the search world as a training variable, using frontier diagnosis to locate the units of a task space where the policy is neither saturated nor collapsed, and synchronized support injection to grow new tasks there with the evidence they need. \textit{iii)} We let the current policy decide how each new task is used, admitting for RL the candidates whose probes disagree and turning those beyond its reach into anchor verified demonstrations of its own, so no external model enters the loop. \textit{iv)} Experiments at two scales show gains of \hyp{intro.d9}{9.8} and \hyp{intro.d27}{9.5} average points over the recipe it builds on, of which \hyp{intro.directed}{7.0} are out of reach for a merely consistent world.

\begin{figure*}[!t]
    \centering
    \IfFileExists{images/main-figure.pdf}{%
        \includegraphics[width=1.0\linewidth]{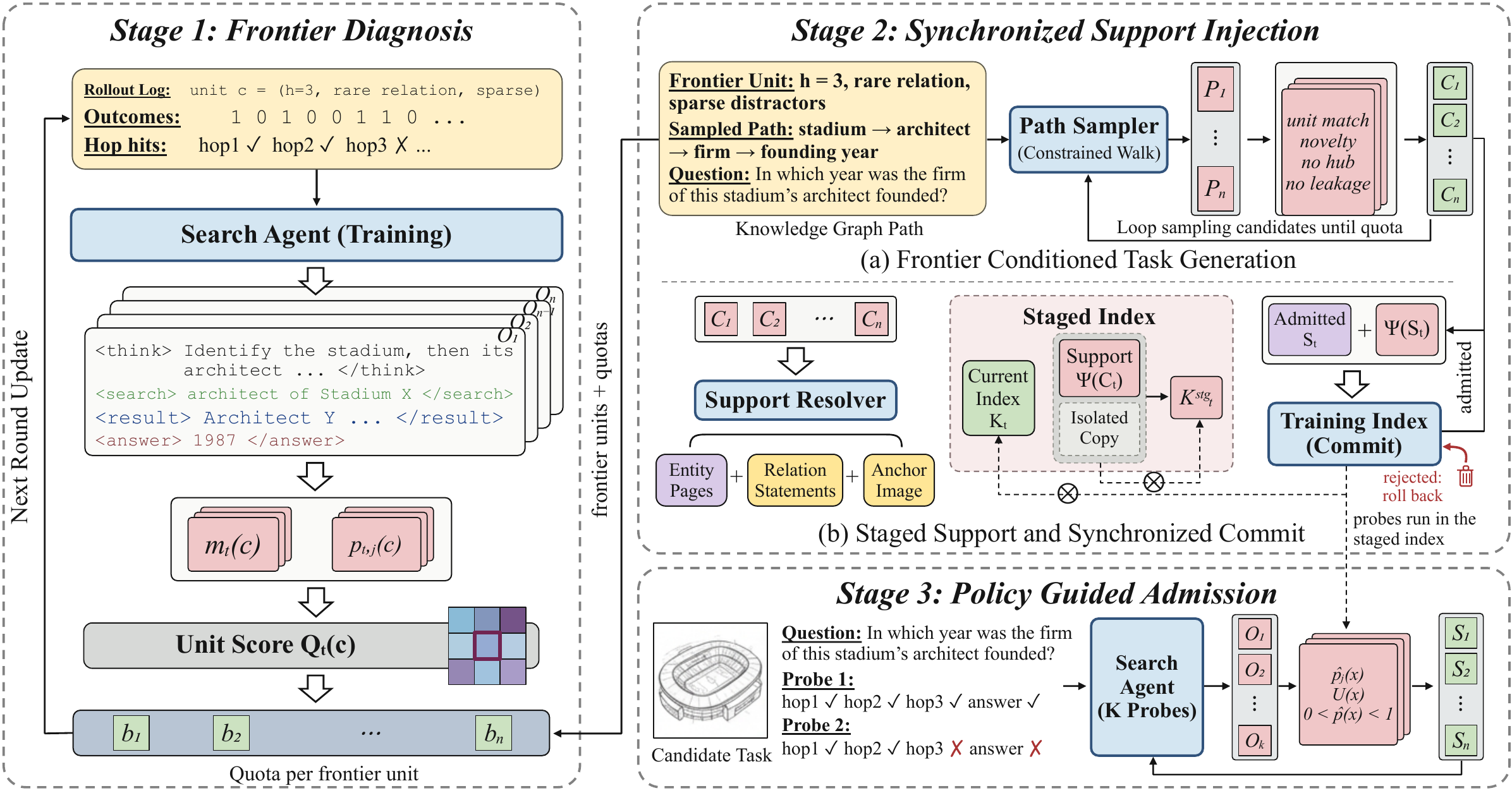}%
    }{%
        \fbox{\parbox[c][3.7in][c]{0.97\linewidth}{\centering
        \textbf{Stage 1: Frontier Diagnosis} (left, full height) \qquad
        \textbf{Stage 2: Synchronized Support Injection} (top right) \qquad
        \textbf{Stage 3: Policy Guided Admission and Self Demonstration} (bottom right)}}%
    }
    \vspace{-4mm}
    \caption{\textbf{Overview of \ours.} We first diagnose the capability frontier from the rollout logs of the current policy $\pi_t$, yielding a quota of new tasks for each partially mastered unit (Stage 1). We then generate candidates in these units and stage their evidence in an isolated copy of the training index (Stage 2). The same policy probes every candidate: candidates on which the probes disagree are admitted for RL, those it fails entirely are retried under a privileged hint and kept as verified demonstrations, and only the evidence of kept candidates is committed (Stage 3). The trained $\pi_{t+1}$ runs the next round, closing the recursive loop.}
    \vspace{-2mm}
    \label{fig:main_figure}
\end{figure*}

\section{Related Work}
\paragraph{Multimodal Search Agents.}
Retrieval augmented generation \cite{lewis2020retrieval,asai2024self} and tool using agents \cite{yao2022react} have been extended to images, from knowledge intensive VQA \cite{marino2019ok,wang2017fvqa,chen2023can} to open web search and long horizon browsing \cite{jiang2024mmsearch,cheng2025simplevqa,fu2026seeking,zeng2026vision,zhang2026browsecomp,zhang2026visbrowse}. Agents built on MLLMs \cite{liu2023visual,bai2025qwen3} learn retrieval actions with RL, covering multimodal search \cite{wu2025mmsearch}, visual evidence seeking \cite{geng2026webwatcher,huang2026vision,zhang2026visual}, high resolution perception \cite{chng2026sensesearch}, and long horizon research \cite{zhang2026deepvoyager,tang2026long}. Open recipes improve the data \cite{chen2026opensearch,jiao2026searcheyes,li2026onesearch,huang2026towards}, the rollout allocation \cite{dai2026simplesearch}, or the optimizer \cite{shao2024deepseekmath,yu2026dapo,gao2025soft}. However, all of them keep the content of the training index fixed, so whether a task is answerable is decided once, when the index is built. In contrast, \ours makes the index a variable that grows with the task pool and is complementary to these recipes.

\paragraph{Curricula and Self Improvement.}
Curricula prioritize learnable tasks \cite{bengio2009curriculum,portelas2020teacher,jiang2021prioritized}, and adaptive medical agents refine visual reasoning and diagnostic workflows \cite{zhu2026medeyes,zhu2026path,zhu2026medsynapse}. Self improvement supplies training signals through correctness filtered rationales \cite{zelikman2022star,singh2023beyond}, alternating SFT and RL on model rollouts \cite{zhang2025evolvesearch}, teachers conditioned on privileged context \cite{zhao2026self}, or self proposed tasks \cite{zhao2026absolute,huang2026r,lu2026search,he2026visplay}. Agent frameworks co evolve environments with the agent, often with an external model that diagnoses weaknesses and writes tasks \cite{dong2026agent,fan2026environment}, and recent \textit{recursive} self improvement feeds the improved model back into the mechanism that improves it, at the skill \cite{zhang2026rsipo} or the inference level \cite{lu2026arex}. These methods assume that every proposed task is solvable in principle, which does not hold in search, where a task is learnable only if its evidence is retrievable. \ours places the recursion at the level of the training world, so the world must expand to keep the self generated tasks \textit{answerable}.

\section{Preliminaries}
\subsection{Search Agents and Search Worlds}
Search agents play a central role in grounding MLLM outputs in external evidence. An agent is an MLLM parameterized as a policy $\pi_{\theta}$. Given a task $x=(I,q,y)$ with an image $I$, a question $q$, and a verifiable answer $y$, it interleaves reasoning with tool calls (visual search, text search, page lookup, summarization, and a Python interpreter) and finally submits an answer. We consider tasks synthesized from a typed multimodal knowledge graph $\mathcal{G}$, each rendered from a supporting path $\pi_x=(e_0\xrightarrow{r_1}e_1\cdots\xrightarrow{r_h}e_h)$ whose anchor $e_0$ is shown in the image and whose last node $e_h$ is the answer.
We define the \textit{search world} at round $t$ as $\searchworld_t=(\taskpool_t,\knowledge_t)$, where $\taskpool_t$ is the pool of trainable tasks and $\knowledge_t$ the evidence exposed by the training index; the tool interface is fixed. We write $\Psi(x)$ for the \textit{support set} of $x$, \textit{i.e.}, the anchor image, entity pages, and verbalized relation statements on $\pi_x$ that must be retrievable for the path to be recoverable; a triple of $\mathcal{G}$ need not be stated in any indexed document. Static recipes keep both parts fixed, $\searchworld_t=(\taskpool_0,\knowledge_0)$; since synthesis is offline, tasks with missing support can be dropped, so the pool is consistent with the index, but the agent soon saturates on it.

Search agents are commonly trained in two stages \cite{jiao2026searcheyes,li2026onesearch,tang2026long}: a supervised warm up on filtered expert trajectories, followed by RL. For a task $x$, the policy samples a group of $G$ trajectories, and each trajectory receives a format reward $r_{\mathrm{format}}$ and an answer reward $r_{\mathrm{answer}}$, which is 1 when the submitted answer matches $y$ and 0 otherwise. Based on these verifiable rewards, the agent optimizes
\begin{eqnarray}
    \mathcal{L}_{\mathrm{RL}} &=& - \mathbb{E}_{x\sim\taskpool_t,\, \tau\sim \pi_{\theta}(\cdot|x;\knowledge_t)} [r(x,\tau)] \nonumber \\ && + \beta\, \mathbb{D}_{\mathrm{KL}}(\pi_{\theta}\,\|\,\pi_{\theta_{\mathrm{old}}}),
    \label{eq:rl}
\end{eqnarray}
where $r=r_{\mathrm{format}}+r_{\mathrm{answer}}$ and $\mathbb{D}_{\mathrm{KL}}(\cdot)$ denotes the KL divergence. In practice, we adopt the hop anchored optimizer of \textsc{SearchEyes} \cite{jiao2026searcheyes}, a variant of GRPO \cite{shao2024deepseekmath} (see the appendix). Note that the rollouts of Eq.~(\ref{eq:rl}) are executed in $\knowledge_t$, which is where the limitations below originate.

\subsection{Limitations of Online Task Growth}
\paragraph{Unreachable Negative Problem.}
Online recipes enlarge the pool every round while leaving the index untouched:
\begin{eqnarray}
\taskpool_{t+1} \ = \ \taskpool_t\cup\admitset_t, \qquad \knowledge_{t+1} \ = \ \knowledge_t, \label{eq:static-growth}
\end{eqnarray}
where $\admitset_t$ is the set of $B$ newly admitted tasks. A task is \textit{reachable} if $\Psi(x)\subseteq\knowledge_t$. When a single element of $\Psi(x)$ is missing, the outcome is $0$ for every policy, and we call the resulting failure an \textit{unreachable negative}. To measure how much of a growing pool remains learnable, we track the \textit{environment consistency}
\begin{eqnarray}
\Cons_t & = & \frac{1}{|\taskpool_t|}\sum_{x\in\taskpool_t}\mathbb{1}\big[\Psi(x)\subseteq\knowledge_t\big]. \label{eq:consistency}
\end{eqnarray}
Under Eq.~(\ref{eq:static-growth}), $\Cons_t$ decays with every round, and group normalization turns the zeros of unreachable tasks into penalties on trajectories that may already be correct. Note that reachability guarantees that evidence \textit{exists}, not that the agent \textit{retrieves} it; retrieval and reasoning errors on reachable tasks remain part of what the policy must learn.

\paragraph{Frontier Mismatch Problem.}
Growth schedules decide \textit{how many} tasks to add but rarely \textit{where}. A scalar success rate cannot tell a unit that fails at one specific hop from one that fails uniformly, so new tasks land in saturated or collapsed regions where the sampled groups contain a single outcome and the advantages are close to zero.

\paragraph{Zero Signal Problem.}
Even for a reachable task, Eq.~(\ref{eq:rl}) yields no gradient when every rollout fails, because all advantages in the group vanish. The tasks just beyond the frontier, which online growth is meant to supply, are therefore invisible to RL, and filtering them out \cite{yu2026dapo} only hides the problem.

\section{\ours}
In this section, we introduce \textbf{\ours}, which is designed to scale online task growth by expanding the search world along the agent's capability frontier. An overview is shown in Fig.~\ref{fig:main_figure}. Each round first diagnoses where the agent is partially competent, then grows tasks in those regions together with their evidence, and finally lets the current policy decide how each new task enters the training of the next policy. The three steps address the three limitations above in turn, and the appendix gives the full algorithm of one round.

\subsection{Frontier Diagnosis}
To overcome the coarseness of scalar difficulty, \ours begins by describing every task with a unit $c=(h,r,d)\in\unitspace$ of hop depth, relation rarity, and distractor density, discretized into $3{\times}3{\times}3$ cells (listed in the appendix), since tasks with the same success rate fail for unrelated reasons along these axes: a late hop that breaks, a relation whose query is malformed, or a look alike entity.
Specifically, after each round we read two statistics per unit from the logged rollouts on reachable tasks: the mastery $m_t(c)$, \textit{i.e.}, the success rate of unit $c$, and the hop hit rate $p_{t,j}(c)$, \textit{i.e.}, the fraction of rollouts that recover the path node $e_j$. Following the view that a unit is useful only when its rollouts disagree, we score each unit by
\begin{equation}
\score_t(c) = \berv\big[m_t(c)\big] + \tfrac{1}{h}\textstyle\sum_{j=1}^{h}\berv\big[p_{t,j}(c)\big]. \label{eq:qscore}
\end{equation}
Here $\berv[x]=x(1-x)$ is the variance of a Bernoulli variable of mean $x$, so both terms lie in $[0,\frac{1}{4}]$ and we keep them unweighted, which a sweep in the appendix supports: the first vanishes for saturated or collapsed units, and the second separates a unit that fails at one specific hop from one that fails uniformly. The frontier $\frontier_t$ keeps the units whose mastery lies in a band $[\tau_\ell,\tau_u]$ and has not dropped since the last round, since a unit whose mastery decreases is being forgotten rather than learned and more tasks there would amplify the drift. Each frontier unit receives a quota $b_t(c)\propto\score_t(c)$ under a fixed budget $B=\sum_c b_t(c)$.
Note that the budget is identical for all compared methods, so the diagnosis changes \textit{where} tasks are added, never \textit{how many} of them enter training.

\begin{table*}[!t]
\centering
\caption{\textbf{Results on nine multimodal search benchmarks} (VisSearch in Table~\ref{tab:diagnostic_res}). Accuracy (\%) judged by GPT-5. Avg.\ is the mean over the six standard benchmarks (\textsc{SimpleVQA} to \textsc{FVQA}). Superscripts give the source of reported values: $^{\text{a}}$\citet{jiao2026searcheyes}, $^{\text{b}}$\citet{li2026onesearch}, $^{\text{c}}$\citet{tang2026long}. $^\dagger$ marks controlled reruns sharing the backbone, supervised checkpoint, optimizer, tools, task budget, and judge with \ours, and $^\ddagger$ marks cells that the source does not report and that we evaluated ourselves under the same protocol. Conclusions rest on the $^\dagger$ block; other rows differ in tools and budgets and give context.}
\label{tab:main_res}
\begin{adjustbox}{width=\textwidth,center}
\input{tables/main_res}
\end{adjustbox}
\vspace{-3mm}
\end{table*}

\subsection{Synchronized Support Injection}
Ideally, once the frontier is known, new tasks in those units should directly strengthen the agent. In practice, however, this is hindered by two barriers: a \textit{task gap} (the generator may emit tasks that do not match the requested unit or that leak their answer) and an \textit{evidence gap} (the index may not contain what the new tasks need).
Addressing these gaps, \ours grows each task together with its evidence and commits it in step with training.

\paragraph{Frontier Conditioned Task Generation.}
Following the constrained walks of \citet{chen2026opensearch} and \citet{jiao2026searcheyes}, we sample paths \textit{conditioned on a frontier unit}. For each $c\in\frontier_t$, we draw walks $\gamma\sim\mathcal{W}(\mathcal{G}\mid c)$ and render every accepted walk into an image grounded task $x=\mathcal{R}(\gamma)$ with the \textit{frozen} renderer $\mathcal{R}$ of the initial pool. Writing $u(\gamma)$ for the unit that $\gamma$ realizes, a walk is accepted if and only if
\begin{equation}
\begin{aligned}
&u(\gamma)=c,\quad \gamma\notin\{\pi_x\}_{x\in\taskpool_t},\\
&\textstyle\max_{j<h}\mathrm{deg}^{-}(e_j)\le\tau_{\mathrm{hub}},\quad \mathcal{A}(e_{1:h})\cap q=\varnothing,
\end{aligned}
\label{eq:accept}
\end{equation}
where $\mathrm{deg}^{-}$ is the in degree in $\mathcal{G}$ and $\mathcal{A}(\cdot)$ the alias set of the hidden nodes. These conditions keep the generator on the requested unit, block shortcuts through popular entities, and prevent $q$ from leaking an intermediate answer.
The accepted tasks form the candidate set $\candset_t$, which is about three times larger than the budget so that the next stage has room to select.
Note that the renderer only verbalizes paths that the diagnosis has requested and never judges difficulty or correctness, so it acts as part of the environment rather than as a teacher. Additionally, to prevent world growth from injecting test time evidence, we discard every candidate whose support touches an answer or an entity of the evaluation sets.

\paragraph{Staged Support and Synchronized Commit.}
To address the evidence gap, we introduce a stage then commit mechanism that grows the index with the same granularity as the tasks.
Specifically, given the candidates $\candset_t$, we first resolve their support sets from $\mathcal{G}$ and write them into an isolated copy of the training index:
\begin{eqnarray}
\Kstg_t &=& \knowledge_t\ \cup \textstyle\bigcup_{x\in\candset_t}\Psi(x).
\end{eqnarray}
We then judge every candidate in this staged world (Section~\ref{sec:admission}) and update the world jointly through a two step process: 1) the kept tasks and their support are committed, \textit{i.e.}, $\knowledge_{t+1}=\knowledge_t\cup\bigcup_{x\in\admitset_t\cup\demoset_t}\Psi(x)$; 2) the support of all other candidates is rolled back together with the copy.
Staging is required because a candidate must be judged under the evidence it will actually be trained with: probing in $\knowledge_t$ would mark every unsupported candidate as hopeless, and the long or rare paths behind unreachable negatives would be discarded instead of repaired.
Note that we commit the support of kept tasks only. We aim to keep the index free of facts that no task uses, since such facts enlarge the neighborhoods of look alike entities and silently shift the difficulty of existing units; loading all of $\mathcal{G}$ is the extreme case. Building on this setting, every newly kept task is reachable by construction, and the evaluation index $\Keval$ is a frozen snapshot that never receives staged or committed support. Staging is cheap, since $\knowledge^{\mathrm{stg}}$ is a delta over $\knowledge_t$ that the probes read as a union with it, and rejected support is dropped with its candidate, so the index grows by under \hyp{m.growth}{1}\% in ten rounds. 

\begin{figure*}[!t]
    \centering
    \includegraphics[width=\linewidth]{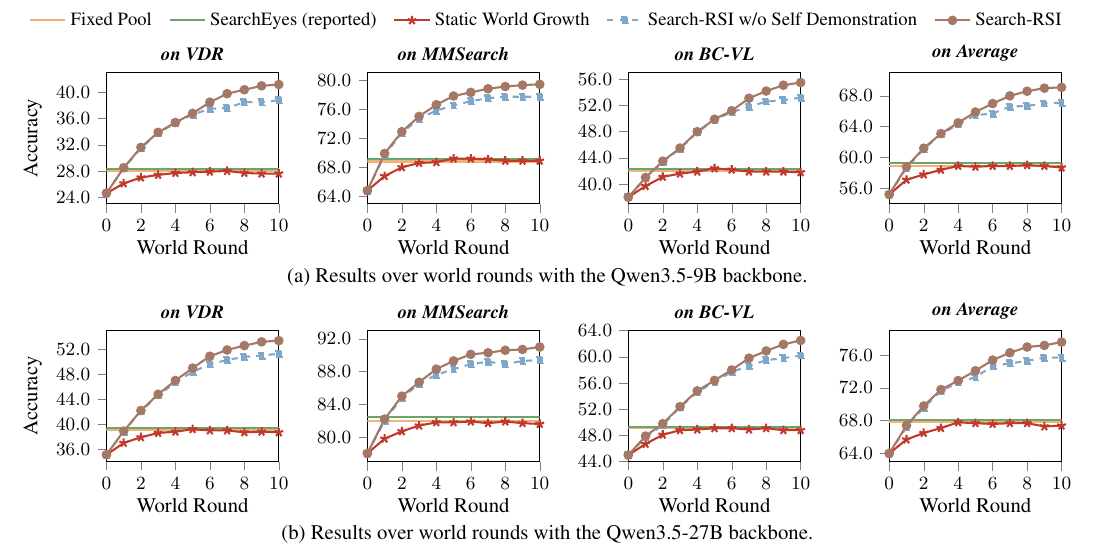}
    \vspace{-7mm}
    \caption{\textbf{Performance over world rounds} on \textsc{VDR}, \textsc{MMSearch}, BC-VL, and the six benchmark average. Static world growth saturates and degrades, while \ours keeps improving and moves away from its RL only variant.}
    \vspace{-3mm}
    \label{fig:scaling_res}
\end{figure*}

\subsection{Policy Guided Admission}
\label{sec:admission}
The goal of this stage is to let the current policy decide how each candidate can still move it. We run $K$ short probe rollouts per candidate in $\Kstg_t$, which are never added to the training buffer. Let $\hat{p}_j(x)$ be the fraction of probes that recover $e_j$ and $\hat{p}(x)$ the fraction that answer correctly. Candidates with $\hat{p}(x)=1$ are already mastered and discarded, while the remaining ones are routed into two disjoint streams.

\paragraph{Admission for RL.}
For candidates with $0<\hat{p}(x)<1$, the group will contain both outcomes. Note that averaging the signed advantages of the probes is not informative, as group centering makes this average zero for every candidate. We therefore rank candidates by their hop level disagreement:
\begin{eqnarray}
U(x) &=& \frac{1}{h}\sum_{j=1}^{h}\hat{p}_j(x)\big(1-\hat{p}_j(x)\big), \label{eq:utility}
\end{eqnarray}
and admit, within each frontier unit, the top $b_t(c)$ of them into $\admitset_t$, so that $\taskpool_{t+1}=\taskpool_t\cup\admitset_t$. A high $U(x)$ means that the probes split at specific hops, which is exactly where the hop anchored optimizer can assign credit.

\paragraph{Self Demonstration.}
Candidates with $\hat{p}(x)=0$ lie beyond the current policy, and RL cannot learn from them because every rollout fails. Inspired by rationalization \cite{zelikman2022star} and privileged self distillation \cite{zhao2026self}, we let the \textit{same} policy retry each such candidate in $\Kstg_t$ up to $M$ times under a privileged hint $\eta(x)$ that lists the relation chain of $\pi_x$ and its nodes. A trajectory is kept only if it is correct and \textit{anchor first}, \textit{i.e.}, every hidden node $e_j$ appears in a tool observation before it appears in any token generated by the agent, so that the hint can direct the search but cannot replace it. We then remove $\eta(x)$ from the context and add the resulting pair $(x,\tilde{\tau})$ to the demonstration set $\demoset_t$, on which a short supervised step precedes the next RL phase:
\begin{equation}
\mathcal{L}_{\mathrm{SD}} = -\mathbb{E}_{(x,\tilde{\tau})\sim\demoset_t}\Big[\frac{1}{|\tilde{\tau}|}\sum_{k}\log\pi_\theta(\tilde{\tau}_k\mid x,\tilde{\tau}_{<k})\Big], \label{eq:sd}
\end{equation}
where $\demoset_t$ collects all kept pairs of round $t$ and observation tokens are masked, so that the policy imitates only its own reasoning and tool calls, never the retrieved text. Demonstrations thus extend learning beyond the frontier, which RL alone cannot reach, while the hint never enters training. The supervised step comes before RL so that some demonstrated tasks acquire mixed outcomes, and hence nonzero gradients, in the next RL phase.

\begin{table*}[!t]
\centering
\caption{\textbf{Search world and frontier diagnostics.} Cons is the final environment consistency and UN the share of unreachable negatives among negative rewards. EM is measured in the frozen index $\Keval$ on the source split, the independent split (also by hop depth $h$), and VisSearch (reported values: \hyp{vs.se9}{18.6} and \hyp{vs.se27}{24.3}). Buckets are assigned by the round 0 checkpoint. \textcolor{darkgreen}{Green}: gains over static growth; \textcolor{red}{red}: its loss of consistency. Cons and UN diagnose the mechanism; EM measures accuracy.}
\label{tab:diagnostic_res}
\vspace{-2mm}
\begin{adjustbox}{width=\textwidth,center}
\input{tables/diagnostic_res}
\end{adjustbox}
\vspace{-3mm}
\end{table*}

\paragraph{Why the Loop Is Recursive.}
We call a loop recursive when the retained improvement reaches the mechanism that produces the next improvement \cite{zhang2026rsipo,lu2026arex}. In \ours, the improver at round $t$ is the policy $\pi_t$ itself: it supplies the statistics of Eq.~(\ref{eq:qscore}), the probes of Eq.~(\ref{eq:utility}), and the demonstrations of Eq.~(\ref{eq:sd}), whereas the source graph, the path sampler, and the frozen renderer stay unchanged. A better $\pi_{t+1}$ therefore diagnoses a different frontier, admits different tasks, and demonstrates tasks that $\pi_t$ could not. This recursion is bounded by the coverage of $\mathcal{G}$, and we test it directly by freezing the improver at $\pi_0$ (Section~\ref{sec:ablation}).

\begin{figure*}[!t]
    \vspace{2mm}
    \centering
    \IfFileExists{images/case-study.pdf}{%
        \includegraphics[width=\linewidth]{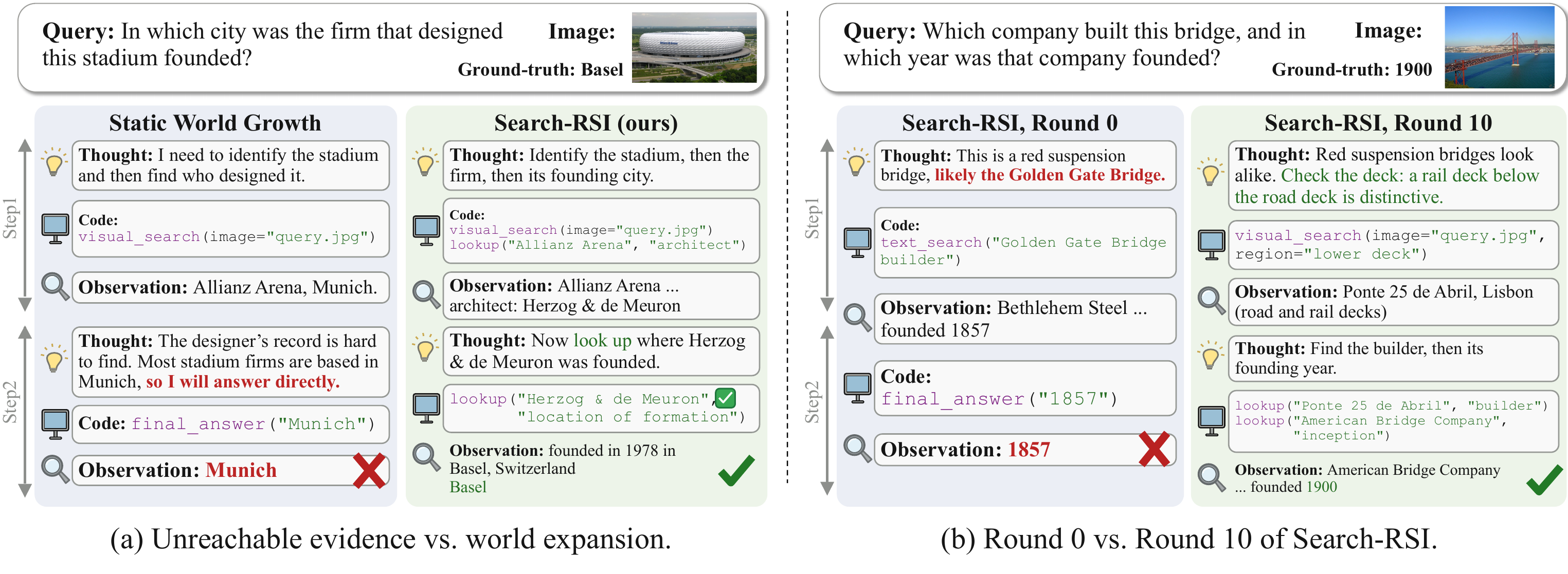}%
    }{\fbox{\parbox[c][3.4cm][c]{0.96\linewidth}{\centering\small Figure 5 placeholder: qualitative comparison.}}}
    \vspace{-4mm}
    \caption{\textbf{Qualitative comparison.} (a) Static world growth answers \textit{Munich} from memory, whereas \ours retrieves the architect and returns \textit{Basel}. (b) The round 0 policy mistakes the bridge for the Golden Gate Bridge, whereas the round 10 policy identifies the Ponte 25 de Abril and resolves its builder.}
    \label{fig:case}
    \vspace{-4mm}
\end{figure*}

\section{Experiments}

\paragraph{Implementation Details.}
We train Qwen3.5-9B and 27B \cite{team2026qwen3} on 16 H20 GPUs and reproduce the supervised stage of \textsc{SearchEyes} \cite{jiao2026searcheyes} with its reported recipe, so every compared run starts from the same checkpoint. We build on its graph, knowledge base, and synthesis pipeline, and take its \hyp{setup.d0}{12k} RL questions as $\taskpool_0$, so round 0 coincides with it. Each run trains for \hyp{setup.steps}{200} RL steps with group size 8 and adds \hyp{setup.B}{1{,}240} tasks every \hyp{setup.tw}{20} steps over \hyp{setup.rounds}{10} rounds, so all pools have identical size; \ours adds \hyp{setup.demos}{4.0k} self demonstrations. We set $[\tau_\ell,\tau_u]=[0.2,0.8]$, $K{=}4$, $M{=}4$, and run one epoch of Eq.~(\ref{eq:sd}) per round with an equal number of replayed warm up trajectories.

\begin{figure}[!t]
\centering
\includegraphics[width=0.96\columnwidth]{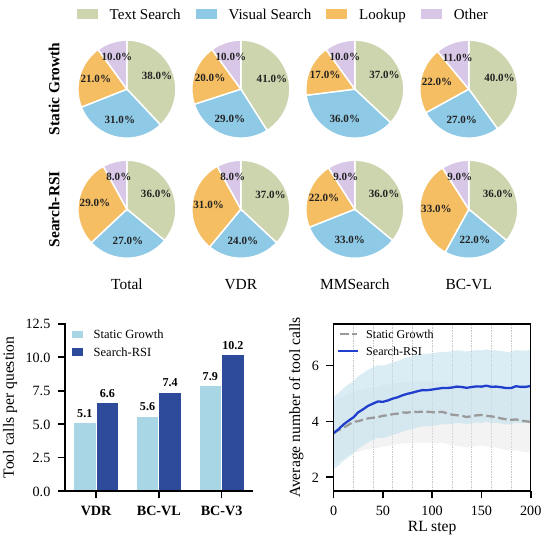}
\vspace{-4mm}
\caption{\textbf{Tool use analysis (9B).} Top: distribution of tool calls, Other merging summarization and Python. Bottom: mean calls per question on long chains, and per training trajectory (shaded: one std; dotted: world expansions).}
\label{fig:tooluse}
\vspace{-4mm}
\end{figure}
\paragraph{Baselines.}
Our primary baseline is \textit{Static World Growth}, which adds the same number of tasks online under Eq.~(\ref{eq:static-growth}), uniformly over units and retained at random. \textit{Fixed Pool} reruns the \textsc{SearchEyes} RL stage on a pool of the same final size synthesized offline, and \textit{EvolveSearch} \cite{zhang2025evolvesearch} alternates SFT on its own successful rollouts with RL on a fixed pool.

\paragraph{Benchmarks.}
Following \citet{jiao2026searcheyes}, we evaluate on \textsc{SimpleVQA} \cite{cheng2025simplevqa}, \textsc{VDR} \cite{zeng2026vision}, \textsc{MMSearch} \cite{jiang2024mmsearch}, \textsc{LiveVQA} \cite{fu2026seeking}, BC-VL \cite{geng2026webwatcher}, and \textsc{FVQA} \cite{wang2017fvqa}, and add InfoSeek \cite{chen2023can}, BC-V3 \cite{zhang2026browsecomp}, VisBrowse \cite{zhang2026visbrowse}, and the in world VisSearch \cite{jiao2026searcheyes}. Web benchmarks use the search API of \textsc{SearchEyes}; VisSearch and two held out splits of \hyp{setup.diag}{2k} tasks, a \textit{source} split from the training partition of $\mathcal{G}$ and an \textit{independent} split from a disjoint one, use the frozen index $\Keval$.

\subsection{Results}
\paragraph{Multimodal Search Benchmarks.}
We draw conclusions only from the controlled block of Table~\ref{tab:main_res}, where \ours outperforms every baseline at both scales. With the same number of RL tasks, it surpasses static world growth by $\mathbf{\hyp{res.d9}{+10.4}}$ and $\mathbf{\hyp{res.d27}{+10.2}}$ on the average and EvolveSearch by \hyp{res.evo9}{+9.7} and \hyp{res.evo27}{+9.4}, so the gain comes from \textit{what} enters training, not from more data. Static world growth even falls below the fixed pool (\hyp{res.s27}{67.4} vs.\ \hyp{res.f27}{67.8} at 27B): fresh tasks do not help when many are unreachable. The largest gains appear on the longest chains, BC-V3 ($\mathbf{\hyp{res.bcv3}{+14.0}}$), \textsc{VDR}, and BC-VL. Agents trained on the live web \cite{tang2026long,zhang2026deepvoyager} remain ahead on \textsc{LiveVQA} and BC-VL, which reward pages outside any snapshot; the two directions may combine, which we have not tested. Fig.~\ref{fig:scaling_res} traces the comparison over rounds.

\paragraph{Search World and Frontier.}
Table~\ref{tab:diagnostic_res} shows that \ours removes the failure it targets: static growth drives consistency to \hyp{diag.cs}{0.85} with \hyp{diag.un}{41.0}\% unreachable negatives, while \ours keeps it at \hyp{diag.co}{0.98}. The gains grow with hop depth (\hyp{diag.h2}{+5.6} at $h{=}2$ vs.\ \hyp{diag.h4}{+13.3} at $h{=}4$ for 9B), concentrate on frontier (\hyp{diag.dfront}{+13.3}) and beyond frontier tasks (\hyp{diag.dbeyond}{+15.9}), and transfer to the independent split and to VisSearch (\hyp{diag.vs.s}{17.9}$\rightarrow$\hyp{diag.vs.o}{28.6}), whose entities were never written to the training index. EvolveSearch, which self trains in a fixed world, improves Beyond by only \hyp{diag.evo.b}{1.2} over the fixed pool.

\subsection{Ablation Studies}
\label{sec:ablation}

\paragraph{RQ0: Is a simpler repair of the world enough?}
The appendix further compares simpler repair strategies at equal budget. Accepting only walks with indexed support ensures consistency, but unsupported paths are often the deepest and rarest, so the frontier share drops to \hyp{rq0.scg.front}{24.8}\% and the average stops at \hyp{rq0.scg}{60.9}. Repairing evidence afterwards reaches \hyp{rq0.rand}{62.1} and an index holding the entire graph reaches \hyp{rq0.oracle}{63.9}, still \hyp{rq0.oraclegap}{5.2} points behind: the \hyp{rq0.total}{10.4} point gain splits into \hyp{rq0.part1}{3.4} points from consistency and \hyp{rq0.part2}{7.0} points from directed growth. 

\paragraph{RQ1: What is the effect of each component?}
Fig.~\ref{fig:case} shows the failure modes it removes. In Table~\ref{tab:ablation_study}, support injection contributes most (\hyp{abl.dssi}{-3.6}), and most of the Beyond gain disappears without it. Frontier diagnosis and guided admission each add \hyp{abl.dpga}{2.3} to \hyp{abl.dfd}{2.6} at unchanged consistency, and self demonstration costs \hyp{abl.dsd}{2.0} on the average but \hyp{abl.dsd.b}{5.8} on Beyond. All gaps hold over three seeds under a paired bootstrap with Holm correction.

\begin{table}[!t]
\centering
\caption{\textbf{Component ablation (9B).} Front.\ is the share of admitted tasks in frontier units and Beyond the EM on beyond frontier tasks. Avg.\ changes are relative to the full model. Frozen Improver keeps $\pi_0$ for diagnosis, probes, and demonstrations. Means of three seeds.}
\label{tab:ablation_study}
\begin{adjustbox}{width=\columnwidth,center}
\input{tables/ablation_study}
\end{adjustbox}
\vspace{-3mm}
\end{table}

\paragraph{RQ2: Is the self improvement recursive?}
Freezing the improver at $\pi_0$ costs \hyp{abl.frozen}{3.2} on the average and \hyp{abl.frozen.b}{4.9} on Beyond, and its demonstration yield stays at \hyp{abl.yield.f}{18}\% while that of \ours reaches \hyp{abl.yield1}{41}\%. We read this as \textit{policy in the loop} world expansion rather than open ended recursion: an external 27B teacher changes the result by at most \hyp{abl.teacher}{0.1}, so what matters is that the improver tracks the policy being trained, not that it grows stronger. Self demonstration in a static world or search self play \cite{lu2026search} recovers less than a fifth of the gain, so the loop needs both the improved policy and the expanding world; the appendix separates the two. 

\paragraph{RQ3: How does search behavior change?}
Fig.~\ref{fig:tooluse} shows that under static world growth, tool calls per trajectory peak at step \hyp{tool.peak}{80} and then decline, since a correct search on an unreachable task is penalized, so it answers early. \ours instead extends its search (\hyp{tool.s}{4.0}$\rightarrow$\hyp{tool.o}{5.3} calls, \hyp{tool.bcv3.s}{7.9}$\rightarrow$\hyp{tool.bcv3.o}{10.2} on BC-V3), and the added calls are mostly page lookups (\hyp{tool.lu.s}{21}\%$\rightarrow$\hyp{tool.lu.o}{29}\%), the hops that injection repairs.

\paragraph{RQ4: Does \ours scale across world rounds?}
Fig.~\ref{fig:scaling_res} shows that static world growth tracks \ours in the first rounds and then saturates and degrades, as each round adds more unreachable negatives. \ours keeps improving, crosses both reference lines after two rounds, and moves away from its RL only variant as the improved policy demonstrates more tasks. Its own curve flattens over the last rounds, as the frontier units exhaust the paths that $\mathcal{G}$ supports. The trend holds at both scales, so online growth helps mainly when the world expands with it.

\section{Conclusion}
We introduced \ours, a recursive self improvement framework that expands the search world along the capability frontier. Online growth helps mainly when the evidence grows with the tasks and the improved policy routes them; frontier diagnosis targets learnable units, support injection removes unreachable negatives, and guided admission with self demonstration reaches past them without a teacher. The loop needs an improver aligned with the policy, not a stronger one; live web environments remain future work.

\clearpage
{
    \small
    \bibliographystyle{ieeenat_fullname}
    \bibliography{main,missing_refs}
}

\clearpage
\appendix
\section*{Supplementary Material}
\input{appendix}

\end{document}

%% file: preamble.tex
\usepackage{cuted} 

\usepackage{currfile} 

\usepackage{caption} 



%% file: tables/main_res.tex
\renewcommand{\arraystretch}{1.05}
\fontsize{7pt}{8pt}\selectfont
\setlength{\tabcolsep}{1.0mm}
\begin{tabular}{lccccccccc@{\hspace{2.5mm}}c}
\toprule
\textbf{Model} & \textbf{SimpleVQA} & \textbf{VDR} & \textbf{MMSearch} & \textbf{LiveVQA} & \textbf{BC-VL} & \textbf{FVQA} & \textbf{InfoSeek} & \textbf{BC-V3} & \textbf{VisBrowse} & \textbf{Avg.} \\
\midrule
\rowcolor{gray!15}\multicolumn{11}{c}{\textit{Agentic Workflows with Proprietary MLLMs}} \\
Gemini-3.1-Pro$^{\text{a}}$~\cite{google2026gemini31} & 80.5$^\ddagger$ & 41.2$^\ddagger$ & 86.1 & 76.6 & 64.1 & 84.0 & 78.4$^\ddagger$ & 52.1$^\ddagger$ & 63.0$^\ddagger$ & 72.1$^\ddagger$ \\ 
GPT-5.5$^{\text{c}}$~\cite{openai2026gpt55} & 82.3 & 43.5$^\ddagger$ & 82.7 & 90.3 & 68.2 & 85.2$^\ddagger$ & 79.6$^\ddagger$ & 55.0 & 66.3 & 75.4$^\ddagger$ \\ 
\midrule
\rowcolor{gray!15}\multicolumn{11}{c}{\textit{Open Multimodal Search Agents}} \\
SenseSearch-8B$^{\text{a,b}}$~\cite{chng2026sensesearch} & 61.7 & 19.4 & 67.4 & 56.2 & 35.1 & 67.1 & 61.7 & 18.2$^\ddagger$ & 33.0$^\ddagger$ & 51.2 \\ 
OpenSearch-VL-8B$^{\text{a,b}}$~\cite{chen2026opensearch} & 71.6 & 20.8 & 64.5 & 59.6 & 37.6 & 71.5 & 70.2 & 20.4$^\ddagger$ & 35.1$^\ddagger$ & 54.3 \\ 
OneSearch-VL-8B$^{\text{b}}$~\cite{li2026onesearch} & 72.4 & 23.6 & 66.5 & 61.2 & 39.0 & 73.4 & 72.3 & 22.6$^\ddagger$ & 36.4$^\ddagger$ & 56.0 \\ 
SimpleSearch-VL-8B$^{\text{c}}$~\cite{dai2026simplesearch} & 76.6 & 25.1$^\ddagger$ & 77.1 & 75.2 & 52.1 & 74.8$^\ddagger$ & 73.5$^\ddagger$ & 29.8$^\ddagger$ & 44.0$^\ddagger$ & 63.5$^\ddagger$ \\ 
DeepVoyager-VL-8B$^{\text{c}}$~\cite{zhang2026deepvoyager} & 76.3 & 26.0$^\ddagger$ & 72.7 & 82.7 & 58.4 & 75.3$^\ddagger$ & 73.9$^\ddagger$ & 32.3 & 47.3 & 65.2$^\ddagger$ \\ 
Long-MDR-9B$^{\text{c}}$~\cite{tang2026long} & 79.3 & 29.4$^\ddagger$ & 74.3 & 83.1 & 59.1 & 76.0$^\ddagger$ & 74.6$^\ddagger$ & 35.7 & 49.6 & 66.9$^\ddagger$ \\ 
OpenSearch-VL-32B$^{\text{a}}$~\cite{chen2026opensearch} & 76.2 & 33.8 & 72.3 & 70.5 & 43.8 & 74.7 & 74.8$^\ddagger$ & 26.5$^\ddagger$ & 41.2$^\ddagger$ & 61.9 \\ 
SimpleSearch-VL-30B-A3B$^{\text{c}}$~\cite{dai2026simplesearch} & 79.6 & 30.6$^\ddagger$ & 83.6 & 81.1 & 55.9 & 77.2$^\ddagger$ & 76.1$^\ddagger$ & 33.1$^\ddagger$ & 48.5$^\ddagger$ & 68.0$^\ddagger$ \\ 
DeepVoyager-VL-30B-A3B$^{\text{c}}$~\cite{zhang2026deepvoyager} & 81.0 & 31.5$^\ddagger$ & 74.0 & 82.7 & 64.2 & 77.8$^\ddagger$ & 76.5$^\ddagger$ & 35.0 & 53.8 & 68.5$^\ddagger$ \\ 
\midrule
\rowcolor{gray!15}\multicolumn{11}{c}{\textit{Online Task Growth on the SearchEyes Recipe}} \\
Qwen3.5-9B (agentic)$^{\text{a}}$~\cite{team2026qwen3} & 56.8 & 19.3 & 41.6 & 55.4 & 31.7 & 63.2 & 58.4$^\ddagger$ & 11.6$^\ddagger$ & 21.3$^\ddagger$ & 44.7 \\ 
\hspace{3mm} + SearchEyes$^{\text{a}}$~\cite{jiao2026searcheyes} & 75.4 & 28.3 & 69.2 & 66.1 & 42.3 & 74.2 & 70.9$^\ddagger$ & 24.1$^\ddagger$ & 37.8$^\ddagger$ & 59.3 \\ 
\hspace{3mm} + Fixed Pool$^\dagger$ & 75.1 & 28.0 & 68.8 & 65.8 & 42.0 & 73.9 & 70.8 & 24.2 & 37.9 & 58.9 \\ 
\hspace{3mm} + Static World Growth$^\dagger$ & 74.9 & 27.6 & 68.9 & 65.9 & 41.8 & 72.9 & 70.5 & 23.8 & 37.5 & 58.7 \\ 
\hspace{3mm} + EvolveSearch$^\dagger$~\cite{zhang2025evolvesearch} & 75.6 & 28.5 & 69.4 & 66.0 & 42.6 & 74.1 & 71.2 & 25.1 & 38.4 & 59.4 \\ 
\rowcolor{PastaYellow} \hspace{3mm} + \textbf{Search-RSI (Ours)} & \textbf{81.9} & \textbf{41.2} & \textbf{79.4} & \textbf{74.6} & \textbf{55.5} & \textbf{82.0} & \textbf{78.6} & \textbf{37.8} & \textbf{51.2} & \textbf{69.1} \\ 
\midrule
Qwen3.5-27B (agentic)$^{\text{a}}$~\cite{team2026qwen3} & 64.3 & 25.8 & 59.1 & 50.6 & 39.4 & 67.3 & 64.0$^\ddagger$ & 16.2$^\ddagger$ & 27.4$^\ddagger$ & 51.1 \\ 
\hspace{3mm} + SearchEyes$^{\text{a}}$~\cite{jiao2026searcheyes} & 80.9 & 39.4 & 82.4 & 77.3 & 49.3 & 79.1 & 76.4$^\ddagger$ & 29.4$^\ddagger$ & 42.7$^\ddagger$ & 68.1 \\ 
\hspace{3mm} + Fixed Pool$^\dagger$ & 80.6 & 39.0 & 82.0 & 77.0 & 49.1 & 78.8 & 75.5 & 30.8 & 43.7 & 67.8 \\ 
\hspace{3mm} + Static World Growth$^\dagger$ & 80.2 & 38.7 & 81.6 & 76.6 & 48.8 & 78.6 & 75.1 & 30.1 & 43.2 & 67.4 \\ 
\hspace{3mm} + EvolveSearch$^\dagger$~\cite{zhang2025evolvesearch} & 81.1 & 39.6 & 82.5 & 77.4 & 49.6 & 79.2 & 76.0 & 31.8 & 44.4 & 68.2 \\ 
\rowcolor{PastaYellow} \hspace{3mm} + \textbf{Search-RSI (Ours)} & \textbf{87.6} & \textbf{53.4} & \textbf{91.0} & \textbf{85.6} & \textbf{62.5} & \textbf{85.5} & \textbf{82.9} & \textbf{44.6} & \textbf{57.3} & \textbf{77.6} \\ 
\bottomrule
\end{tabular}

%% file: tables/diagnostic_res.tex
\renewcommand{\arraystretch}{1.2}
\fontsize{7pt}{8pt}\selectfont
\setlength{\tabcolsep}{0.75mm}
\begin{tabular}{lcc ccc ccc ccc}
\toprule
& \multicolumn{2}{c}{\textbf{Search World}} & \multicolumn{3}{c}{\textbf{In World EM}} & \multicolumn{3}{c}{\textbf{Indep.\ EM by Hop Depth}} & \multicolumn{3}{c}{\textbf{EM by Competence Bucket}} \\
\cmidrule(lr){2-3} \cmidrule(lr){4-6} \cmidrule(lr){7-9} \cmidrule(lr){10-12}
\textbf{Method} & \textbf{Cons}$\uparrow$ & \textbf{UN (\%)}$\downarrow$ & \textbf{Source} & \textbf{Indep.} & \textbf{VisSearch} & $h{=}2$ & $h{=}3$ & $h{=}4$ & \textbf{Saturated} & \textbf{Frontier} & \textbf{Beyond} \\
\midrule
\rowcolor{gray!15}\multicolumn{12}{c}{\textit{Qwen3.5-9B Backbone}} \\
Round 0 checkpoint & 0.95 & 3.1 & 69.1 & 64.2 & 14.6 & 80.1 & 63.5 & 49.0 & 84.0 & 51.6 & 29.8 \\
\hspace{3mm} + Fixed Pool & 0.95 & 3.1 & 72.4 & 69.6 & 18.3 & 83.9 & 69.4 & 55.5 & 86.6 & 62.5 & 38.9 \\
\hspace{3mm} + Static World Growth & \textcolor{red}{0.85} & \textcolor{red}{41.0} & 72.0 & 69.2 & 17.9 & 83.8 & 69.1 & 54.7 & 86.4 & 63.1 & 38.7 \\
\hspace{3mm} + EvolveSearch & 0.95 & \textbf{3.0} & 73.0 & 70.2 & 18.9 & 84.1 & 70.0 & 56.5 & 86.8 & 64.0 & 40.1 \\
\rowcolor{PastaYellow} \hspace{3mm} + \textbf{Search-RSI (Ours)} & \textbf{0.98} & 4.4 & \textbf{82.4} $\scriptstyle\textcolor{darkgreen}{(\textbf{+10.4})}$ & \textbf{79.6} $\scriptstyle\textcolor{darkgreen}{(\textbf{+10.4})}$ & \textbf{28.6} $\scriptstyle\textcolor{darkgreen}{(\textbf{+10.7})}$ & \textbf{89.4} $\scriptstyle\textcolor{darkgreen}{(\textbf{+5.6})}$ & \textbf{81.3} $\scriptstyle\textcolor{darkgreen}{(\textbf{+12.2})}$ & \textbf{68.0} $\scriptstyle\textcolor{darkgreen}{(\textbf{+13.3})}$ & \textbf{90.8} $\scriptstyle\textcolor{darkgreen}{(\textbf{+4.4})}$ & \textbf{76.4} $\scriptstyle\textcolor{darkgreen}{(\textbf{+13.3})}$ & \textbf{54.6} $\scriptstyle\textcolor{darkgreen}{(\textbf{+15.9})}$ \\
\midrule
\rowcolor{gray!15}\multicolumn{12}{c}{\textit{Qwen3.5-27B Backbone}} \\
Round 0 checkpoint & 0.95 & 2.9 & 75.0 & 72.0 & 20.1 & 86.5 & 71.6 & 57.9 & 88.3 & 60.2 & 37.0 \\
\hspace{3mm} + Fixed Pool & 0.95 & 2.9 & 80.1 & 77.2 & 24.0 & 89.4 & 77.6 & 64.6 & 90.8 & 70.4 & 46.1 \\
\hspace{3mm} + Static World Growth & \textcolor{red}{0.85} & \textcolor{red}{39.6} & 79.8 & 76.9 & 23.5 & 89.3 & 77.3 & 64.1 & 90.6 & 70.9 & 45.8 \\
\hspace{3mm} + EvolveSearch & 0.95 & \textbf{2.8} & 80.6 & 77.7 & 24.6 & 89.6 & 78.0 & 65.5 & 90.9 & 71.2 & 47.0 \\
\rowcolor{PastaYellow} \hspace{3mm} + \textbf{Search-RSI (Ours)} & \textbf{0.98} & 4.1 & \textbf{89.9} $\scriptstyle\textcolor{darkgreen}{(\textbf{+10.1})}$ & \textbf{86.1} $\scriptstyle\textcolor{darkgreen}{(\textbf{+9.2})}$ & \textbf{34.0} $\scriptstyle\textcolor{darkgreen}{(\textbf{+10.5})}$ & \textbf{93.4} $\scriptstyle\textcolor{darkgreen}{(\textbf{+4.1})}$ & \textbf{87.6} $\scriptstyle\textcolor{darkgreen}{(\textbf{+10.3})}$ & \textbf{77.2} $\scriptstyle\textcolor{darkgreen}{(\textbf{+13.1})}$ & \textbf{94.0} $\scriptstyle\textcolor{darkgreen}{(\textbf{+3.4})}$ & \textbf{83.1} $\scriptstyle\textcolor{darkgreen}{(\textbf{+12.2})}$ & \textbf{61.3} $\scriptstyle\textcolor{darkgreen}{(\textbf{+15.5})}$ \\
\bottomrule
\end{tabular}

%% file: tables/ablation_study.tex
\renewcommand{\arraystretch}{1.15}
\fontsize{7pt}{8pt}\selectfont
\setlength{\tabcolsep}{1.4mm}
\begin{tabular}{lccccc l}
\toprule
\textbf{Configuration} & \textbf{Cons} & \textbf{UN} & \textbf{Front.} & \textbf{Beyond} & \textbf{Indep.} & \textbf{Avg.} \\
\midrule
Static World Growth & 0.85 & 41.0 & 31.5 & 38.7 & 69.2 & 58.7 $\scriptstyle\textcolor{red}{(\textbf{-10.4})}$ \\
\midrule
\rowcolor{gray!15}\multicolumn{7}{c}{\textit{Component Ablation}} \\
\textit{w/o} Frontier Diagnosis & 0.98 & 4.6 & 33.0 & 51.8 & 77.9 & 66.5 $\scriptstyle\textcolor{red}{(\textbf{-2.6})}$ \\
\textit{w/o} Support Injection & 0.93 & 8.6 & 36.8 & 48.9 & 76.0 & 65.5 $\scriptstyle\textcolor{red}{(\textbf{-3.6})}$ \\
\textit{w/o} Guided Admission & 0.98 & 4.5 & 43.1 & 52.4 & 78.2 & 66.8 $\scriptstyle\textcolor{red}{(\textbf{-2.3})}$ \\
\textit{w/o} Self Demonstration & 0.98 & 4.5 & \textbf{44.2} & 48.8 & 78.3 & 67.1 $\scriptstyle\textcolor{red}{(\textbf{-2.0})}$ \\
\textit{w/o} Hop Ambiguity in $\mathcal{Q}_t$ & 0.98 & 4.5 & 42.0 & 53.1 & 78.8 & 67.8 $\scriptstyle\textcolor{red}{(\textbf{-1.3})}$ \\
\midrule
\rowcolor{gray!15}\multicolumn{7}{c}{\textit{Recursion and Alternative Sources of Improvement}} \\
Frozen Improver ($\pi_0$) & 0.98 & 4.6 & 37.5 & 49.7 & 76.6 & 65.9 $\scriptstyle\textcolor{red}{(\textbf{-3.2})}$ \\
Static World + Self Demos. & 0.85 & 39.8 & 31.5 & 42.0 & 71.0 & 60.1 $\scriptstyle\textcolor{red}{(\textbf{-9.0})}$ \\
Search Self Play$^\dagger$~\cite{lu2026search} & 0.96 & 5.2 & 35.2 & 42.8 & 71.8 & 60.6 $\scriptstyle\textcolor{red}{(\textbf{-8.5})}$ \\
External Teacher Demos. & 0.98 & 4.4 & 44.0 & \textbf{54.9} & \textbf{79.8} & \textbf{69.2} $\scriptstyle\textcolor{darkgreen}{(\textbf{+0.1})}$ \\
\midrule
\rowcolor{PastaYellow} \textbf{Search-RSI (Full)} & \textbf{0.98} & \textbf{4.4} & \textbf{44.2} & 54.6 & 79.6 & \textbf{69.1} \\
\bottomrule
\end{tabular}

%% file: appendix.tex
\renewcommand{\thesection}{\Alph{section}}
\renewcommand{\thetable}{S\arabic{table}}
\renewcommand{\thefigure}{S\arabic{figure}}
\renewcommand{\thealgorithm}{S\arabic{algorithm}}
\setcounter{section}{0}\setcounter{table}{0}\setcounter{figure}{0}\setcounter{algorithm}{0}
\renewcommand{\theHtable}{appendix.\arabic{table}}
\renewcommand{\theHfigure}{appendix.\arabic{figure}}
\providecommand{\theHalgorithm}{\thealgorithm}
\renewcommand{\theHalgorithm}{appendix.\arabic{algorithm}}

\section{More Implementation Details}
\label{app:impl}

\subsection{Training Process}
\label{app:rl}
Each world round of \ours trains the policy with two objectives in sequence: a short supervised step on the self demonstrations $\demoset_t$ (Eq.~(\ref{eq:sd})) and $T_w$ RL steps on the pool $\taskpool_{t+1}$ inside $\knowledge_{t+1}$ (Eq.~(\ref{eq:rl})). The two streams are kept apart: $\taskpool_{t+1}=\taskpool_t\cup\admitset_t$ receives exactly $B$ admitted tasks per round, and a demonstrated task is used only by the supervised step. Its support is still committed, so that the task stays reachable when it is probed again as a candidate in a later round and can then be admitted for RL through the usual route; the \hyp{app.data.demos2}{4.0k} demonstrations are therefore not extra RL tasks. The supervised step comes first so that the RL phase starts from a policy that has already seen a successful route through the hardest committed tasks, which turns some of them into tasks with mixed outcomes and hence nonzero gradients.

\paragraph{Policy Optimization.}
For a task $x$, the policy samples a group of $G$ trajectories with rewards $\{r_i\}_{i=1}^{G}$. GRPO \cite{shao2024deepseekmath} normalizes the rewards inside the group,
\begin{equation}
\hat{A}_i=\frac{r_i-\mathrm{mean}(\{r_j\}_{j=1}^{G})}{\mathrm{std}(\{r_j\}_{j=1}^{G})+\delta},
\end{equation}
and maximizes the clipped surrogate
\begin{equation}
\mathcal{J}(\theta)=\mathbb{E}\Big[\frac{1}{|\tau_i|}\sum_{k}\min\big(\rho_{i,k}\hat{A}_i,\ \mathrm{clip}(\rho_{i,k},1{-}\epsilon,1{+}\epsilon)\hat{A}_i\big)\Big],
\end{equation}
where $\rho_{i,k}=\pi_\theta(\tau_{i,k}\mid x,\tau_{i,<k})/\pi_{\theta_{\mathrm{old}}}(\tau_{i,k}\mid x,\tau_{i,<k})$ is the token importance ratio and observation tokens are masked. The hop anchored optimizer of \textsc{SearchEyes} \cite{jiao2026searcheyes}, which we use by default, mixes $\hat{A}_i$ with hop level advantages and gates negative updates asymmetrically. \ours only reads the per hop outcomes of the rollouts and adds a supervised step between rounds, so the optimizer is used without modification, and GRPO or \textsc{SAPO} \cite{gao2025soft} can be swapped in (Table~\ref{tab:optimizer}).

\paragraph{Self Demonstration.}
For a candidate with $\hat{p}(x)=0$, the policy is prompted with the question, the image, and the hint $\eta(x)$ of Appendix~\ref{app:prompts}, and acts in the staged index $\Kstg_t$ with the normal tool interface. A trajectory passes the anchor first check if, for every hidden node $e_j$ on $\pi_x$, the first occurrence of the name or an alias of $e_j$ in the trajectory lies inside a tool observation rather than inside a token generated by the agent. Trajectories that refer to the hint (\textit{e.g.}, ``as the hint says'') are removed by string matching. At most two passing trajectories are kept per task, after which $\eta(x)$ is deleted from the prompt, so the stored pair has exactly the format of an ordinary training trajectory.

\subsection{Implementation Details}
\textbf{Compute.}
All experiments run on a single node of 16 NVIDIA H20 GPUs. This includes the supervised warm up, every round of world expansion (probing, hinted retries, and the supervised step on self demonstrations), and RL for both the 9B and the 27B backbone. The same hardware serves all compared methods, so the relative GPU hours in Table~\ref{tab:cost} are measured under identical conditions.

\textbf{Supervised Warm Up.}
We reproduce the supervised stage of \textsc{SearchEyes} with its reported recipe; Table~\ref{tab:sft_params} lists the hyperparameters. The same checkpoint initializes every compared run.

\begin{table}[!t]
\centering
\caption{Key hyperparameters of the supervised warm up.}
\label{tab:sft_params}
\vspace{-2mm}
\resizebox{0.86\linewidth}{!}{
\begin{tabular}{@{}lc@{}}
\toprule
\textbf{Hyperparameter} & \textbf{Value} \\ \midrule
Base model & Qwen3.5-9B or 27B \cite{team2026qwen3} \\
Questions / trajectories & 10K / 30K \cite{jiao2026searcheyes} \\
Learning rate & 2e-5 \\
Epochs & 3 \\
LR scheduler & Cosine \\
Fine tuning & Full parameters \\
Batch size & \hyp{app.sft.bs}{64} \\ 
Cutoff length & \hyp{app.sft.len}{32k} tokens \\ 
\bottomrule
\end{tabular}}
\vspace{-2mm}
\end{table}

\textbf{RL and World Expansion.}
Table~\ref{tab:rl_params} lists the RL and world expansion hyperparameters. All values not listed are the \textsc{SearchEyes} defaults, and all compared methods share them.

\begin{table}[!t]
\centering
\caption{Key hyperparameters of RL and world expansion.}
\label{tab:rl_params}
\vspace{-2mm}
\resizebox{0.86\linewidth}{!}{
\begin{tabular}{@{}lc@{}}
\toprule
\textbf{Hyperparameter} & \textbf{Value} \\ \midrule
Optimizer & HaPO \cite{jiao2026searcheyes} \\
RL steps & 200 \\
Group size $G$ & 8 \\
KL coefficient $\beta$ & 0.02 \\
Hop advantage mixing ratio & 0.3 \\
Gate temperatures (pos. / neg.) & 1.0 / 1.05 \\
World rounds / interval $T_w$ & \hyp{app.rounds}{10} / \hyp{app.tw}{20} steps \\ 
Budget per round $B$ & \hyp{app.B}{1{,}240} \\ 
Mastery band $[\tau_\ell,\tau_u]$ & $[0.2, 0.8]$ \\
Probes per candidate $K$ & 4 \\
Hub threshold $\tau_{\mathrm{hub}}$ & \hyp{app.hub}{500} in links \\ 
Max tool calls per trajectory & \hyp{app.turns}{30} \\ 
\bottomrule
\end{tabular}}
\vspace{-2mm}
\end{table}

\textbf{Self Demonstration.}
Table~\ref{tab:sd_params} lists the hyperparameters of the self demonstration step. The replayed warm up trajectories are sampled from the existing 30K set and are not regenerated, so no model other than the current policy produces training data during the loop.

\begin{table}[!t]
\centering
\caption{Key hyperparameters of self demonstration.}
\label{tab:sd_params}
\vspace{-2mm}
\resizebox{0.86\linewidth}{!}{
\begin{tabular}{@{}lc@{}}
\toprule
\textbf{Hyperparameter} & \textbf{Value} \\ \midrule
Hinted retries $M$ & 4 \\
Kept trajectories per task & $\leq 2$ \\
Filter & Correct answer, anchor first \\
Learning rate & \hyp{app.sd.lr}{5e-6} \\ 
Epochs per round & 1 \\
Replay ratio (demonstrations : warm up) & 1 : 1 \\
\bottomrule
\end{tabular}}
\vspace{-2mm}
\end{table}

\textbf{Infrastructure.}
Training uses the multi turn rollout engine of \textsc{SearchEyes} with \texttt{vLLM} serving. The training index $\knowledge_t$, the staged copy $\Kstg_t$, and the frozen evaluation index $\Keval$ are separate instances of the \textsc{SearchEyes} retrieval service, so staging and rollback never touch the evaluation backend. Web benchmarks are evaluated with the same search API as \textsc{SearchEyes}.

\subsection{Task Units and Path Constraints}
\label{app:units}
Table~\ref{tab:task_unit} describes the three axes of the unit space. Hop depth takes values $h\in\{2,3,4\}$. Relation rarity is split into three bins at its training set tertiles, and distractor density into three bins at $\{\hyp{app.d1}{2},\hyp{app.d2}{6}\}$ competing entities, where two entities compete if they share a type and the cosine similarity of their CLIP \cite{radford2021learning} image embeddings exceeds $\hyp{app.kappa}{0.85}$. As the one exception to the band rule of the main paper, units with fewer than \hyp{app.nmin}{16} logged rollouts are treated as frontier units, since their mastery cannot be estimated yet and new units would otherwise never be explored. The leakage check matches the question against the name and all aliases of the hidden nodes after lowercasing and lemmatization, and the novelty check compares the ordered relation and node signature of the path and the perceptual hash of the anchor image against the current pool. 

\begin{table*}[!t]
    \centering
    \caption{Description of the structured task units used for frontier diagnosis.}
    \vspace{-2mm}
    \resizebox{\linewidth}{!}{
    \input{tables/task_unit_description}}
    \vspace{-3mm}
    \label{tab:task_unit}
\end{table*}

\subsection{Dataset Details}
\label{app:data}
\textbf{Training Data.}
Table~\ref{tab:data_statistics} summarizes the data consumed by a full run. The warm up and the initial pool are the two disjoint splits of \textsc{SearchEyes}, and every later task is grown by \ours or by the compared protocol under an identical task budget per round.

\begin{table}[!t]
\centering
\caption{Training data statistics of a full run.}
\label{tab:data_statistics}
\vspace{-2mm}
\resizebox{\linewidth}{!}{
\begin{tabular}{@{}l|c|c@{}}
\toprule
\textbf{Split} & \textbf{Size} & \textbf{Producer} \\ \midrule
Warm up questions / trajectories & 10K / 30K & \textsc{SearchEyes} \\
Initial RL pool $\taskpool_0$ & 12K & \textsc{SearchEyes} \\
RL tasks added ($10\times B$) & \hyp{app.data.add}{12.4K} & Growth protocol \\ 
Final RL pool $\taskpool_{10}$ & \hyp{app.data.final}{24.4K} & -- \\ 
Self demonstrations $\cup_t\demoset_t$ (supervised step only) & \hyp{app.data.demos}{4.0K} & Current policy \\ 
Source / independent world splits & 2K / 2K & Held out \\
\bottomrule
\end{tabular}}
\vspace{-2mm}
\end{table}

\textbf{Evaluation Benchmarks.}
Table~\ref{tab:bench} lists the ten evaluation benchmarks. We use the evaluation subsets of \citet{jiao2026searcheyes} for the first six, of \citet{li2026onesearch} for InfoSeek, and of \citet{tang2026long} for BC-V3 and VisBrowse.

\begin{table}[!t]
\centering
\caption{Evaluation benchmarks and their retrieval backend.}
\label{tab:bench}
\vspace{-2mm}
\resizebox{\linewidth}{!}{
\begin{tabular}{@{}l|l|c@{}}
\toprule
\textbf{Benchmark} & \textbf{Focus} & \textbf{Backend} \\ \midrule
SimpleVQA \cite{cheng2025simplevqa} & Factual visual QA & Web \\
VDR \cite{zeng2026vision} & Visual deep research & Web \\
MMSearch \cite{jiang2024mmsearch} & Multimodal web search & Web \\
LiveVQA \cite{fu2026seeking} & Recent visual knowledge & Web \\
BC-VL \cite{geng2026webwatcher} & Long horizon browsing & Web \\
FVQA \cite{wang2017fvqa} & Fact based visual QA & Web \\
InfoSeek \cite{chen2023can} & Entity knowledge seeking & Web \\
BC-V3 \cite{zhang2026browsecomp} & Vertical browsing & Web \\
VisBrowse \cite{zhang2026visbrowse} & Visual native browsing, 169 items & Web \\
VisSearch \cite{jiao2026searcheyes} & Multi hop in world, 1{,}000 items & $\Keval$ \\
\bottomrule
\end{tabular}}
\vspace{-2mm}
\end{table}

\subsection{Overall Algorithm}
Algorithm~\ref{alg:search_rsi} summarizes one run of \ours.

\begin{algorithm}[h]
\caption{Training loop of \ours}
\label{alg:search_rsi}
\small
\begin{algorithmic}[1]
\Require world $\searchworld_0=(\taskpool_0,\knowledge_0)$, warm up policy $\pi_0$, graph $\mathcal{G}$, frozen renderer, frozen $\Keval$, budget $B$
\For{$t=0,1,\dots,T-1$}
  \State \textbf{Frontier Diagnosis}: compute $\score_t(c)$ (Eq.~\ref{eq:qscore}) from the logs of $\pi_t$; select $\frontier_t$ and quotas $b_t(c)$
  \State \textbf{Task Generation}: sample paths per frontier unit, render them, apply the four checks $\Rightarrow\candset_t$; drop candidates that touch $\Keval$
  \State \textbf{Staged Support}: $\Kstg_t\gets\knowledge_t\cup\bigcup_{x\in\candset_t}\Psi(x)$
  \State \textbf{Probing}: $K$ rollouts of $\pi_t$ per candidate in $\Kstg_t$
  \State \textbf{Admission}: $\admitset_t\gets$ top $b_t(c)$ by $U(x)$ (Eq.~\ref{eq:utility}) among $0<\hat{p}(x)<1$
  \State \textbf{Self Demonstration}: for $\hat{p}(x)=0$, retry with $\eta(x)$; keep anchor first successes $\Rightarrow\demoset_t$
  \State \textbf{Commit}: $\taskpool_{t+1}\gets\taskpool_t\cup\admitset_t$ (demonstrated tasks return to the candidates of later rounds); $\knowledge_{t+1}\gets\knowledge_t\cup\Psi(\admitset_t\cup\demoset_t)$; roll back the rest
  \State \textbf{Train}: one epoch of Eq.~(\ref{eq:sd}) on $\demoset_t$ with replay, then $T_w$ steps of Eq.~(\ref{eq:rl}) in $\knowledge_{t+1}$ $\Rightarrow\pi_{t+1}$
\EndFor
\end{algorithmic}
\end{algorithm}

\section{Prompts}
\label{app:prompts}
In this section, we list the prompts used by \ours. The agent prompt is inherited from \textsc{SearchEyes} and is shown here in a shortened form.

\begin{tcolorbox}[title=Agent System Prompt (shortened),width=\linewidth,breakable,colback=white,colframe=black!55,fonttitle=\small\bfseries]
\begin{small}
\textcolor{deltared}{Instruction}\\
You are a multimodal search agent. Answer the question about the image by searching for evidence step by step. Available tools: \texttt{visual\_search}, \texttt{text\_search}, \texttt{lookup}, \texttt{summarize}, \texttt{python\_interpreter}. Identify the entity in the image before searching for its attributes. Only answer when the complete chain is supported by evidence, and put only the final entity name inside \texttt{<answer></answer>}.\\[2pt]
\textcolor{deltared}{Input}\\
Image: \{image\} \quad Question: \{question\}
\end{small}
\end{tcolorbox}

\begin{tcolorbox}[title=Privileged Hint for Self Demonstration,width=\linewidth,breakable,colback=white,colframe=black!55,fonttitle=\small\bfseries]
\begin{small}
\textcolor{deltared}{Appended to the agent prompt, removed before training}\\
Reference chain (private, do not quote it): \{$e_0$ in the image\} $\xrightarrow{r_1}$ \{$e_1$\} $\xrightarrow{r_2}\cdots\xrightarrow{r_h}$ \{$e_h$\}.\\
Use the reference only to decide what to search for next. You must retrieve every entity of the chain with a tool before you state it. A step that is not supported by a retrieved observation will be rejected.\\[2pt]
\textcolor{deltared}{Kept trajectory}\\
Correct answer; each $e_j$ first appears in an observation.
\end{small}
\end{tcolorbox}

\begin{tcolorbox}[title=Frozen Renderer,width=\linewidth,breakable,colback=white,colframe=black!55,fonttitle=\small\bfseries]
\begin{small}
\textcolor{deltared}{Instruction}\\
Write one natural question about the image whose answer is the last entity of the path. Refer to the entity in the image only as ``this'' plus its type. Do not mention the name or any alias of a hidden entity. Return only the question.\\[2pt]
\textcolor{deltared}{Input}\\
Path: \{$e_0$\} $\xrightarrow{r_1}\cdots\xrightarrow{r_h}$ \{$e_h$\} \quad Anchor type: \{type of $e_0$\}
\end{small}
\end{tcolorbox}

\begin{tcolorbox}[title=Answer Judge,width=\linewidth,breakable,colback=white,colframe=black!55,fonttitle=\small\bfseries]
\begin{small}
You are a precise answer validator. Compare the predicted answer with the set of acceptable answers. Consider semantic equivalence and ignore minor differences in formatting or capitalization. Return only ``True'' if the prediction matches any acceptable answer, otherwise ``False''.\\[2pt]
Question: \{question\} \quad Correct Answers: \{answers\} \quad Predicted Answer: \{prediction\}
\end{small}
\end{tcolorbox}

\section{Additional Experimental Results}
\label{app:analysis}

\subsection{Full Main Results}
Table~\ref{tab:main_full} extends Table~\ref{tab:main_res} with direct reasoning models, further open search agents, and the variant of \ours trained by RL alone. Cells marked $\ddagger$ are not reported by the source and were evaluated by us under the protocol of \textsc{SearchEyes}.

We read the table at two levels of evidence. The controlled block, marked $\dagger$, shares the backbone, warm up checkpoint, optimizer, tools, task budget, and judge with \ours, and every conclusion of the paper rests on it. The other rows use different tools, turn budgets, and in part different judges, and some of their cells are our own evaluations; they place the controlled results in the landscape of current agents but do not support claims of superiority over them. Three observations stand out. First, direct reasoning without tools stays far below every agentic system on \textsc{VDR}, BC-VL, and BC-V3, whose answers are rarely memorized, so these benchmarks measure search rather than recall. Second, the RL only variant already improves on every benchmark at both scales over the RL stage it replaces, which isolates the contribution of a consistent and frontier focused world from that of self demonstration; the latter adds a further \hyp{app.full.sd9}{2.0} on the average at 9B and \hyp{app.full.sd27}{1.9} at 27B, in line with Table~\ref{tab:ablation_study}, and is largest on BC-V3 and VisBrowse (\hyp{app.full.sdbc}{+3.2} each), the benchmarks with the longest chains. Third, agents trained on the live web with longer horizons, such as DeepVoyager-VL and Long-MDR, reach higher scores on \textsc{LiveVQA} and BC-VL, where recent events and open web pages dominate. Since \ours only changes the world in which the agent is trained, it is orthogonal to these recipes, and we expect the two to combine. 

\begin{table*}[!t]
    \centering
    \caption{Full results on nine benchmarks. Superscripts mark the source of reported values as in Table~\ref{tab:main_res}; $\dagger$ marks controlled reruns and $\ddagger$ cells evaluated by us.}
    \vspace{-2mm}
    \resizebox{0.92\linewidth}{!}{
    \begin{tabular}{l|cccccc|c|ccc}
    \toprule
    \textbf{Model} & \textbf{SimpleVQA} & \textbf{VDR} & \textbf{MMSearch} & \textbf{LiveVQA} & \textbf{BC-VL} & \textbf{FVQA} & \textbf{Avg.} & \textbf{InfoSeek} & \textbf{BC-V3} & \textbf{VisBrowse} \\
    \midrule
    \rowcolor{gray!15}\multicolumn{11}{c}{\textit{\textbf{Direct Reasoning}}} \\
    \midrule
    GPT-4o$^{\text{a,b}}$ \cite{hurst2024gpt} & 51.7 & 1.7 & 18.7 & 28.1 & 5.5 & 48.0 & 25.6 & 52.9 & 3.3$^\ddagger$ & 5.2$^\ddagger$ \\ 
    GPT-5$^{\text{a,b}}$ \cite{singh2025openai} & 61.6 & 9.8 & 35.1 & 44.4 & 48.6 & 54.4 & 42.3 & 61.7 & 11.9$^\ddagger$ & 18.7$^\ddagger$ \\ 
    Gemini-2.5-Pro$^{\text{a,b}}$ \cite{comanici2025gemini} & 63.0 & 8.0 & 39.8 & 60.3 & 43.1 & 60.7 & 45.8 & 46.9 & 13.2$^\ddagger$ & 20.7$^\ddagger$ \\ 
    Qwen3-VL-32B$^{\text{a,b}}$ \cite{bai2025qwen3} & 58.0 & 4.1 & 19.8 & 45.5 & 30.8 & 34.1 & 32.1 & 28.8 & 6.3$^\ddagger$ & 9.9$^\ddagger$ \\ 
    \midrule
    \rowcolor{gray!15}\multicolumn{11}{c}{\textit{\textbf{Agentic Workflow}}} \\
    \midrule
    GPT-5$^{\text{a}}$ \cite{singh2025openai} & 67.3 & 17.6 & 62.7 & 61.0$^\ddagger$ & 57.6 & 62.0 & 54.7$^\ddagger$ & 65.5$^\ddagger$ & 20.6$^\ddagger$ & 32.2$^\ddagger$ \\ 
    Gemini-2.5-Pro$^{\text{a}}$ \cite{comanici2025gemini} & 74.3 & 10.0 & 55.7 & 53.1$^\ddagger$ & 42.3 & 65.0 & 50.1$^\ddagger$ & 57.0$^\ddagger$ & 15.6$^\ddagger$ & 24.4$^\ddagger$ \\ 
    Gemini-3.1-Pro$^{\text{a}}$ \cite{google2026gemini31} & 80.5$^\ddagger$ & 41.2$^\ddagger$ & 86.1 & 76.6 & 64.1 & 84.0 & 72.1$^\ddagger$ & 78.4$^\ddagger$ & 52.1$^\ddagger$ & 63.0$^\ddagger$ \\ 
    Claude-4.6-Opus$^{\text{a}}$ \cite{anthropic2026claude} & 80.4$^\ddagger$ & 32.2$^\ddagger$ & 76.2 & 67.4 & 48.3 & 74.5 & 63.2$^\ddagger$ & 75.6$^\ddagger$ & 27.4$^\ddagger$ & 43.0$^\ddagger$ \\ 
    Kimi-K2.5$^{\text{a}}$ \cite{team2026kimi} & 84.6$^\ddagger$ & 35.6$^\ddagger$ & 76.6 & 76.6 & 50.3 & 76.5 & 66.7$^\ddagger$ & 79.5$^\ddagger$ & 30.3$^\ddagger$ & 47.5$^\ddagger$ \\ 
    GPT-5.5$^{\text{c}}$ \cite{openai2026gpt55} & 82.3 & 43.5$^\ddagger$ & 82.7 & 90.3 & 68.2 & 85.2$^\ddagger$ & 75.4$^\ddagger$ & 79.6$^\ddagger$ & 55.0 & 66.3 \\ 
    \midrule
    \rowcolor{gray!15}\multicolumn{11}{c}{\textit{\textbf{Open Multimodal Search Agents (7B to 9B)}}} \\
    \midrule
    MMSearch-R1-7B$^{\text{a,b,c}}$ \cite{wu2025mmsearch} & 57.4 & 2.9 & 53.8 & 48.4 & 20.9 & 58.4 & 40.3 & 55.1 & 4.0 & 14.0$^\ddagger$ \\ 
    WebWatcher-7B$^{\text{a,c}}$ \cite{geng2026webwatcher} & 54.3 & 10.3 & 49.1 & 51.2 & 21.2 & 45.6$^\ddagger$ & 38.6$^\ddagger$ & 43.5$^\ddagger$ & 4.7 & 14.2$^\ddagger$ \\ 
    Qwen3-VL-8B (agentic)$^{\text{a,b}}$ \cite{bai2025qwen3} & 52.0 & 17.0 & 37.4 & 50.6 & 27.9 & 58.7 & 40.6 & 50.3 & 11.0$^\ddagger$ & 17.2$^\ddagger$ \\ 
    SenseSearch-8B$^{\text{a,b}}$ \cite{chng2026sensesearch} & 61.7 & 19.4 & 67.4 & 56.2 & 35.1 & 67.1 & 51.2 & 61.7 & 18.2$^\ddagger$ & 33.0$^\ddagger$ \\ 
    ODE-8B$^{\text{a}}$ \cite{huang2026towards} & 70.3 & 20.4 & 66.0 & 59.1$^\ddagger$ & 41.9 & 64.7 & 53.7$^\ddagger$ & 63.4$^\ddagger$ & 19.3$^\ddagger$ & 30.2$^\ddagger$ \\ 
    OpenSearch-VL-8B$^{\text{a,b}}$ \cite{chen2026opensearch} & 71.6 & 20.8 & 64.5 & 59.6 & 37.6 & 71.5 & 54.3 & 70.2 & 20.4$^\ddagger$ & 35.1$^\ddagger$ \\ 
    OneSearch-VL-8B$^{\text{b}}$ \cite{li2026onesearch} & 72.4 & 23.6 & 66.5 & 61.2 & 39.0 & 73.4 & 56.0 & 72.3 & 22.6$^\ddagger$ & 36.4$^\ddagger$ \\ 
    POINTS-Seeker-8B$^{\text{c}}$ \cite{liu2026points} & 68.8 & 30.6$^\ddagger$ & 70.8 & 77.7 & 44.4 & 77.2$^\ddagger$ & 61.6$^\ddagger$ & 73.7$^\ddagger$ & 26.1$^\ddagger$ & 40.9$^\ddagger$ \\ 
    Vision-DeepResearch-8B$^{\text{c}}$ \cite{huang2026vision} & 79.8$^\ddagger$ & 31.7$^\ddagger$ & 69.6 & 76.7 & 42.6 & 78.5$^\ddagger$ & 63.1$^\ddagger$ & 75.0$^\ddagger$ & 11.7 & 42.3$^\ddagger$ \\ 
    Visual-Seeker-8B$^{\text{c}}$ \cite{zhang2026visual} & 81.8$^\ddagger$ & 33.3$^\ddagger$ & 72.2 & 71.7$^\ddagger$ & 47.6 & 80.5$^\ddagger$ & 64.5$^\ddagger$ & 76.9$^\ddagger$ & 28.3$^\ddagger$ & 34.7 \\ 
    SimpleSearch-VL-8B$^{\text{c}}$ \cite{dai2026simplesearch} & 76.6 & 25.1$^\ddagger$ & 77.1 & 75.2 & 52.1 & 74.8$^\ddagger$ & 63.5$^\ddagger$ & 73.5$^\ddagger$ & 29.8$^\ddagger$ & 44.0$^\ddagger$ \\ 
    DeepVoyager-VL-8B$^{\text{c}}$ \cite{zhang2026deepvoyager} & 76.3 & 26.0$^\ddagger$ & 72.7 & 82.7 & 58.4 & 75.3$^\ddagger$ & 65.2$^\ddagger$ & 73.9$^\ddagger$ & 32.3 & 47.3 \\ 
    Long-MDR-9B$^{\text{c}}$ \cite{tang2026long} & 79.3 & 29.4$^\ddagger$ & 74.3 & 83.1 & 59.1 & 76.0$^\ddagger$ & 66.9$^\ddagger$ & 74.6$^\ddagger$ & 35.7 & 49.6 \\ 
    \midrule
    \rowcolor{gray!15}\multicolumn{11}{c}{\textit{\textbf{Open Multimodal Search Agents (27B to 32B)}}} \\
    \midrule
    Qwen3-VL-30B (agentic)$^{\text{a}}$ \cite{bai2025qwen3} & 55.1 & 20.2 & 44.2 & 62.0 & 34.1 & 63.0 & 46.4 & 55.3$^\ddagger$ & 14.6$^\ddagger$ & 23.0$^\ddagger$ \\ 
    ODE-30B$^{\text{a}}$ \cite{huang2026towards} & 71.0 & 26.4 & 69.7 & 64.9$^\ddagger$ & 46.1 & 69.3 & 57.9$^\ddagger$ & 69.6$^\ddagger$ & 23.2$^\ddagger$ & 36.4$^\ddagger$ \\ 
    OpenSearch-VL-30B-A3B$^{\text{a}}$ \cite{chen2026opensearch} & 74.9 & 33.5 & 68.7 & 67.4 & 41.1 & 73.2 & 59.8 & 72.6$^\ddagger$ & 25.3$^\ddagger$ & 39.7$^\ddagger$ \\ 
    WebWatcher-32B$^{\text{a,c}}$ \cite{geng2026webwatcher} & 59.0 & 17.1$^\ddagger$ & 55.3 & 58.7 & 27.0 & 57.7$^\ddagger$ & 45.8$^\ddagger$ & 55.1$^\ddagger$ & 8.7 & 22.8$^\ddagger$ \\ 
    OpenSearch-VL-32B$^{\text{a}}$ \cite{chen2026opensearch} & 76.2 & 33.8 & 72.3 & 70.5 & 43.8 & 74.7 & 61.9 & 74.8$^\ddagger$ & 26.5$^\ddagger$ & 41.2$^\ddagger$ \\ 
    Vision-DeepResearch-30B-A3B$^{\text{c}}$ \cite{huang2026vision} & 86.7$^\ddagger$ & 37.4$^\ddagger$ & 69.6 & 77.6 & 53.7 & 85.3$^\ddagger$ & 68.4$^\ddagger$ & 81.5$^\ddagger$ & 31.9$^\ddagger$ & 50.0$^\ddagger$ \\ 
    SimpleSearch-VL-30B-A3B$^{\text{c}}$ \cite{dai2026simplesearch} & 79.6 & 30.6$^\ddagger$ & 83.6 & 81.1 & 55.9 & 77.2$^\ddagger$ & 68.0$^\ddagger$ & 76.1$^\ddagger$ & 33.1$^\ddagger$ & 48.5$^\ddagger$ \\ 
    DeepVoyager-VL-30B-A3B$^{\text{c}}$ \cite{zhang2026deepvoyager} & 81.0 & 31.5$^\ddagger$ & 74.0 & 82.7 & 64.2 & 77.8$^\ddagger$ & 68.5$^\ddagger$ & 76.5$^\ddagger$ & 35.0 & 53.8 \\ 
    \midrule
    \rowcolor{gray!15}\multicolumn{11}{c}{\textit{\textbf{Qwen3.5-9B Backbone}}} \\
    \midrule
    SearchEyes-9B$^{\text{a}}$ \cite{jiao2026searcheyes} & 75.4 & 28.3 & 69.2 & 66.1 & 42.3 & 74.2 & 59.3 & 70.9$^\ddagger$ & 24.1$^\ddagger$ & 37.8$^\ddagger$ \\ 
    Search-RSI \textit{w/o} Self Demo.$^\dagger$ & 80.3 & 38.8 & 77.7 & 72.9 & 53.2 & 79.7 & 67.1 & 77.2 & 34.6 & 48.0 \\ 
    \rowcolor{PastaYellow}
    \textbf{Search-RSI (Ours)} & \textbf{81.9} & \textbf{41.2} & \textbf{79.4} & \textbf{74.6} & \textbf{55.5} & \textbf{82.0} & \textbf{69.1} & \textbf{78.6} & \textbf{37.8} & \textbf{51.2} \\ 
    \midrule
    \rowcolor{gray!15}\multicolumn{11}{c}{\textit{\textbf{Qwen3.5-27B Backbone}}} \\
    \midrule
    SearchEyes-27B$^{\text{a}}$ \cite{jiao2026searcheyes} & 80.9 & 39.4 & 82.4 & 77.3 & 49.3 & 79.1 & 68.1 & 76.4$^\ddagger$ & 29.4$^\ddagger$ & 42.7$^\ddagger$ \\ 
    Search-RSI \textit{w/o} Self Demo.$^\dagger$ & 85.9 & 51.3 & 89.4 & 83.7 & 60.2 & 83.7 & 75.7 & 81.4 & 41.2 & 54.0 \\ 
    \rowcolor{PastaYellow}
    \textbf{Search-RSI (Ours)} & \textbf{87.6} & \textbf{53.4} & \textbf{91.0} & \textbf{85.6} & \textbf{62.5} & \textbf{85.5} & \textbf{77.6} & \textbf{82.9} & \textbf{44.6} & \textbf{57.3} \\ 
    \bottomrule
    \end{tabular}}
    \vspace{-3mm}
    \label{tab:main_full}
\end{table*}

\subsection{Training Dynamics}
Table~\ref{tab:dynamics} tracks five quantities over a full 9B run. The training reward of \ours stays close to \hyp{app.dyn.r}{0.5}, as expected when the admitted tasks sit at the frontier, whereas the fixed pool saturates and static world growth declines once unreachable negatives accumulate. The response length of static world growth falls in the second half of training, mirroring the early answering in Fig.~\ref{fig:tooluse}. On the independent world split, the RL only variant and the full method separate as training proceeds, from \hyp{app.dyn.gap1}{0.8} points at step 100 to \hyp{app.dyn.gap2}{1.3} at step 200, as more self demonstrations enter training. 

The first four columns together explain why static world growth degrades. Its reward falls as the share of unreachable negatives grows from \hyp{app.dyn.un5}{30.0}\% to \hyp{app.dyn.un10}{41.0}\%; at the same time its responses become shorter, because a trajectory that stops early is penalized no more than one that searches correctly for a missing fact. The fixed pool shows the opposite failure: its reward approaches \hyp{app.dyn.fix}{0.7}, so most groups have identical outcomes and contribute no gradient. \ours avoids both, since the consistency of its world stays near \hyp{app.dyn.cons}{0.98} and the admitted tasks keep the success rate in the informative middle range. Its consistency stays below 1.00 only because of the initial pool: the offline filter of \textsc{SearchEyes} leaves $\taskpool_0$ at a consistency of 0.95, while every task admitted later is reachable, so $\Cons_t=(0.95\cdot 12.0\mathrm{k}+1.24\mathrm{k}\cdot t)/(12.0\mathrm{k}+1.24\mathrm{k}\cdot t)$, which gives 0.97 after five rounds and 0.98 after ten. The share of unreachable negatives follows from the same numbers: since every rollout on an unreachable task fails, UN is at least $1-\Cons_t$ divided by the failure rate, which is why the fixed pool, with lower consistency but a higher success rate, shows more unreachable negatives than \ours. 

\subsection{Routing and Self Demonstration Yield}
\label{app:routing}
\begin{figure}[!t]
\centering
\includegraphics[width=\linewidth]{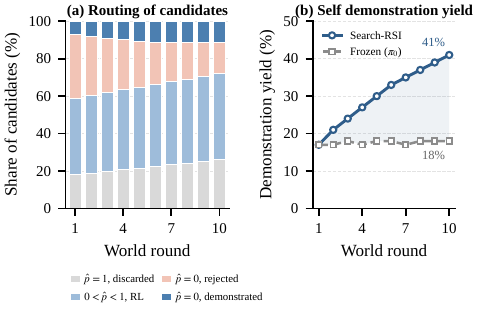}
\vspace{-6mm}
\caption{\textbf{The improver improves (9B).} (a) Share of candidates in each probe route per round. (b) Fraction of failed candidates that the policy turns into verified demonstrations, for \ours and for an improver frozen at $\pi_0$.}
\label{fig:routing}
\vspace{-3mm}
\end{figure}

Fig.~\ref{fig:routing} decomposes the candidates of each round by their probe outcome. About \hyp{app.route.p1}{22}\% are already mastered, \hyp{app.route.mid}{44}\% have mixed outcomes and compete for the RL budget, and \hyp{app.route.p0}{34}\% fail every probe. The yield of self demonstration, \textit{i.e.}, the fraction of failed candidates that yield an anchor first demonstration, rises from \hyp{app.yield0}{17}\% to \hyp{app.yield1}{41}\% as the policy improves, while it stays at about \hyp{app.yield.f}{18}\% when the improver is frozen at $\pi_0$. This growth is the most direct evidence that the improvement of each round reaches the mechanism of the next. 

The composition of the routes also shifts over the run. The share of candidates that are already mastered grows slowly, as the frontier units of earlier rounds become saturated and the diagnosis moves on, while the share that fails every probe shrinks from \hyp{app.route.p0.start}{41}\% in the first round to \hyp{app.route.p0.end}{28}\% in the last. Most of the failed candidates that cannot be demonstrated even with the hint require four hops or a rare relation, and they are kept in the staging area so that they can be probed again once the policy improves. 

\begin{table*}[!t]
\centering
\caption{Training dynamics with the Qwen3.5-9B backbone. Reward, response length (thousand tokens), and exact match on the independent world split are reported at RL steps 100 and 200; environment consistency and the share of unreachable negatives among negative rewards (UN) are reported at world rounds 5 and 10. The fixed pool never changes its world, and the RL only variant shares the world of \ours.}
\label{tab:dynamics}
\vspace{-2mm}
\begin{adjustbox}{width=0.86\textwidth,center}
\fontsize{7pt}{8pt}\selectfont
\renewcommand{\arraystretch}{1.2}
\setlength{\tabcolsep}{2.2mm}
\begin{tabular}{l cc cc cc cc cc}
\toprule
& \multicolumn{2}{c}{\textbf{Train reward}} & \multicolumn{2}{c}{\textbf{Resp. length}} & \multicolumn{2}{c}{\textbf{Cons}$_t$} & \multicolumn{2}{c}{\textbf{UN (\%)}} & \multicolumn{2}{c}{\textbf{Indep. EM}} \\
\cmidrule(lr){2-3}\cmidrule(lr){4-5}\cmidrule(lr){6-7}\cmidrule(lr){8-9}\cmidrule(lr){10-11}
\textbf{Method} & 100 & 200 & 100 & 200 & r5 & r10 & r5 & r10 & 100 & 200 \\
\midrule
Fixed Pool & 0.66 & 0.71 & 2.72 & 2.85 & 0.95 & 0.95 & 12.4 & 13.2 & 68.6 & 69.6 \\
Static World Growth & 0.55 & \textcolor{red}{0.47} & 2.80 & \textcolor{red}{2.55} & 0.88 & \textcolor{red}{0.85} & 30.0 & \textcolor{red}{41.0} & 68.6 & 69.2 \\
\ours \textit{w/o} Self Demo. & 0.52 & 0.53 & 3.07 & 3.32 & 0.97 & 0.98 & 4.3 & 4.4 & 75.4 & 78.3 \\
\rowcolor{PastaYellow}\textbf{\ours} & 0.52 & 0.52 & \textbf{3.13} & \textbf{3.45} & \textbf{0.97} & \textbf{0.98} & \textbf{4.3} & \textbf{4.4} & \textbf{76.2} & \textbf{79.6} \\
\bottomrule
\end{tabular}
\end{adjustbox}
\vspace{-3mm}
\end{table*}

\subsection{Tool Call Distribution}
\begin{figure}[!t]
    \centering
    \includegraphics[width=\linewidth]{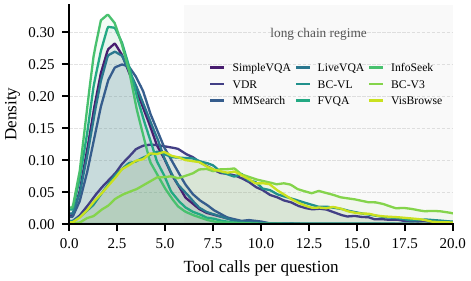}
    \vspace{-6mm}
    \caption{Distribution of tool calls per question for \ours on each benchmark (9B).}
    \vspace{-3mm}
    \label{fig:calls}
\end{figure}

Fig.~\ref{fig:calls} shows the distribution of tool calls per question for \ours on each benchmark. Three regimes emerge. \textbf{(1) Short chains}: \textsc{SimpleVQA}, \textsc{FVQA}, InfoSeek, \textsc{LiveVQA}, and \textsc{MMSearch} concentrate at two to four calls, so most of their questions are answered after identifying the entity and one attribute lookup. \textbf{(2) Long chains}: \textsc{VDR}, BC-VL, and VisBrowse spread over five to twelve calls, consistent with their multi hop design. \textbf{(3) Very long chains}: BC-V3 has the heaviest tail, which matches its larger gain in Table~\ref{tab:main_res}. 

Compared with static world growth, \ours shifts each distribution to the right on the long chain benchmarks but leaves the short chain benchmarks almost unchanged. The extra calls therefore concentrate where the questions need them rather than appearing as a uniform increase in search effort, which would raise the cost without improving accuracy. Within a benchmark, the added calls are mostly page lookups that verify an intermediate entity, a behavior that the per hop breakdown in Table~\ref{tab:diagnostic_res} also reflects. 

\subsection{Selection Rules and Sensitivity}
We also sweep the weight of the second term of Eq.~(\ref{eq:qscore}), writing it as $\score_t(c)=\berv[m_t(c)]+\lambda\,\tfrac{1}{h}\sum_j\berv[p_{t,j}(c)]$, which the main paper sets to $\lambda=1$. The six benchmark average is \hyp{app.lam0}{67.8}, \hyp{app.lam05}{68.8}, \hyp{app.lam2}{69.0}, and \hyp{app.lam4}{68.5} for $\lambda=0$, $0.5$, $2$, and $4$, against \hyp{app.lam1}{69.1} at $\lambda=1$, so the choice is flat in $[0.5,2]$ and falls off outside it: with $\lambda=0$ the diagnosis cannot tell a unit that fails at one hop from one that fails everywhere, and with $\lambda=4$ it chases hop disagreement into units that are not yet learnable, raising the frontier share to \hyp{app.lam4f}{46.1}\% while lowering accuracy. Equal weighting is therefore an empirical choice that the sweep supports, not a consequence of the two terms sharing a range. 

Table~\ref{tab:selection} replaces one selection rule at a time while keeping every other stage and the admitted volume fixed. For the diagnosis, a scalar success band and an absolute learning progress teacher \cite{portelas2020teacher} recover only part of the gain, because they treat units of equal mastery as interchangeable. For admission, the mean signed advantage performs on par with random retention, as expected from group centering, whereas hop level disagreement raises the mean $U$ of the admitted pool from \hyp{app.u0}{0.31} to \hyp{app.u1}{0.43}. Fig.~\ref{fig:sensitivity} further sweeps the mastery band and the probe budget and decomposes the gain over static world growth. 

The mastery band is flat around the default: any band whose lower edge lies between 0.1 and 0.3 and whose upper edge lies between 0.7 and 0.9 stays within \hyp{app.sens.band}{0.6} of the best average, so the method does not depend on a finely tuned threshold. Raising the lower edge to 0.4 hurts more than lowering the upper edge, because it removes units in which the agent has just begun to succeed, which are the units with the largest learning progress. The probe budget saturates at $K=4$: two probes misroute many partially solved tasks to the demonstration route, whereas six or eight probes add cost with almost no gain. The decomposition on the right shows that each component contributes between \hyp{app.sens.min}{2.0} and \hyp{app.sens.max}{3.6} points, with support injection the largest, in line with Table~\ref{tab:ablation_study}. 

\begin{table}[!t]
\centering
\caption{Selection rules at equal admitted volume (9B). Each row replaces one rule while all other stages are kept fixed; colored values give the change over the first row of each block.}
\vspace{-2mm}
\label{tab:selection}
\begin{adjustbox}{width=\linewidth}
\fontsize{7pt}{8pt}\selectfont
\renewcommand{\arraystretch}{1.2}
\setlength{\tabcolsep}{3pt}
\begin{tabular}{l cccc}
\toprule
\textbf{Rule} & \textbf{Source} & \textbf{Indep.} & \textbf{Avg.} & \textbf{Aux.} \\
\midrule
\rowcolor{gray!15}\multicolumn{5}{l}{\textit{Where to grow (Aux.: frontier share of quotas, \%)}} \\
Uniform quotas & 80.4 & 77.9 & 66.5 & 33.0 \\
Absolute learning progress & 80.8 & 78.2 & 67.0 & 36.1 \\
Scalar success band & 81.2 & 78.5 & 67.4 & 39.4 \\
Frontier Diagnosis \textit{w/o} hop ambiguity & 81.6 & 78.8 & 67.8 & 42.0 \\
\rowcolor{PastaYellow}Frontier Diagnosis (Ours) & \textbf{82.4}$\scriptstyle\textcolor{darkgreen}{(\textbf{+2.0})}$ & \textbf{79.6}$\scriptstyle\textcolor{darkgreen}{(\textbf{+1.7})}$ & \textbf{69.1}$\scriptstyle\textcolor{darkgreen}{(\textbf{+2.6})}$ & \textbf{44.2} \\
\midrule
\rowcolor{gray!15}\multicolumn{5}{l}{\textit{What to keep (Aux.: mean hop disagreement $U$ of admitted pool)}} \\
Random retention & 80.6 & 78.1 & 66.8 & 0.31 \\
Mean signed advantage & 80.5$\scriptstyle\textcolor{red}{(\textbf{$-$0.1})}$ & 78.0$\scriptstyle\textcolor{red}{(\textbf{$-$0.1})}$ & 66.7$\scriptstyle\textcolor{red}{(\textbf{$-$0.1})}$ & 0.30 \\
Outcome variance only & 81.4 & 78.6 & 67.6 & 0.36 \\
\rowcolor{PastaYellow}Hop disagreement (Ours) & \textbf{82.4}$\scriptstyle\textcolor{darkgreen}{(\textbf{+1.8})}$ & \textbf{79.6}$\scriptstyle\textcolor{darkgreen}{(\textbf{+1.5})}$ & \textbf{69.1}$\scriptstyle\textcolor{darkgreen}{(\textbf{+2.3})}$ & \textbf{0.43} \\
\bottomrule
\end{tabular}
\end{adjustbox}
\vspace{-3mm}
\end{table}

\subsection{Optimizer Transfer and Controls}
Table~\ref{tab:optimizer} applies \ours to three optimizers without modification. The gain on the independent world split is about \hyp{app.opt}{+10.3} under each of them, so world expansion is additive to improvements in policy optimization. Table~\ref{tab:control} targets five alternative explanations. Training static world growth for $1.27\times$ as many steps, which matches the total cost of \ours, adds only 0.1, so the gain is not bought by compute. Committing the support of every staged candidate raises the consistency to 0.99 but lowers the average, because unused facts enlarge the neighborhoods of look alike entities. An oracle index that contains the whole graph reaches a consistency of 1.00 and still trails \ours. Replacing the frozen renderer with a template or with Qwen3.5-9B changes the average by at most 0.2, so the gain does not come from the renderer. Finally, disabling the evaluation screen changes the average by only 0.1. 

\begin{table}[!t]
\centering
\caption{Optimizer transfer (9B), EM on the independent split.}
\label{tab:optimizer}
\vspace{-2mm}
\resizebox{0.9\linewidth}{!}{
\begin{tabular}{l|cc|c}
\toprule
\textbf{Optimizer} & \textbf{Static Growth} & \textbf{Search-RSI} & $\Delta$ \\
\midrule
GRPO \cite{shao2024deepseekmath} & 68.1 & 78.3 & \up{10.2} \\ 
HaPO \cite{jiao2026searcheyes} & 69.2 & 79.6 & \up{10.4} \\ 
SAPO \cite{gao2025soft} & 69.6 & 80.0 & \up{10.4} \\ 
\bottomrule
\end{tabular}}
\vspace{-2mm}
\end{table}

\begin{table}[!t]
\centering
\caption{Controls for alternative explanations (9B, Avg.).}
\label{tab:control}
\vspace{-2mm}
\resizebox{0.9\linewidth}{!}{
\begin{tabular}{l|c|c}
\toprule
\textbf{Protocol} & \textbf{Cons} & \textbf{Avg.} \\
\midrule
Static World Growth & 0.85 & 58.7 \\ 
\quad + $1.27\times$ steps (compute matched) & 0.85 & 58.8 \\ 
Commit all staged support & 0.99 & 67.4 \\ 
Oracle complete index & 1.00 & 63.9 \\ 
Template renderer & 0.98 & 68.9 \\ 
Qwen3.5-9B renderer & 0.98 & 68.9 \\ 
Evaluation screen disabled & 0.98 & 69.2 \\ 
\rowcolor{PastaYellow}
\textbf{Search-RSI (Ours)} & 0.98 & \textbf{69.1} \\ 
\bottomrule
\end{tabular}}
\vspace{-2mm}
\end{table}

\subsection{Candidate Funnel and Cost}
Table~\ref{tab:audit} traces the candidates of a full run. The path constraints remove shortcuts, leakage, and duplicates but barely change the fraction of candidates with missing support, which confirms that unreachability is not a quality defect that filtering could remove. Among the admitted tasks, \hyp{app.fun.rep}{27.1}\% required injected support, and among the self demonstrated tasks \hyp{app.fun.rep.sd}{38.6}\% did, so long and rare paths are over represented exactly where the policy is weakest. Table~\ref{tab:cost} itemizes the overhead: probes and hinted retries dominate the extra cost but are truncated at the target hop depth, and the total overhead is $1.27\times$ GPU hours. Compute is not the only cost. Each round adds \hyp{app.cost.wall}{1.7} hours of wall clock, of which \hyp{app.cost.wall.probe}{64}\% is probing, \hyp{app.cost.wall.res}{21}\% is resolving support sets from $\mathcal{G}$, and the rest is staging and committing the index. The staged copy is a delta rather than a second index, so the peak storage of the retrieval service grows by \hyp{app.cost.store}{0.8}\%, and the number of retrieval calls it serves rises by \hyp{app.cost.calls}{47}\%, which is the binding constraint when the index is shared across runs. The pipeline also carries engineering cost that no number captures: a staging layer with rollback, a resolver from support statements to documents, and a probe scheduler. We report these so that the method can be judged on its operational footprint and not on accuracy alone. 

\begin{table}[!t]
\centering
\caption{Candidate funnel over a full 9B run.}
\label{tab:audit}
\vspace{-2mm}
\resizebox{\linewidth}{!}{
\begin{tabular}{l|c|cc}
\toprule
\textbf{Stage} & \textbf{Count} & \textbf{Unsupported} & \textbf{Duplicate} \\
\midrule
Sampled paths & 62{,}000 & 29.0\% & 11.6\% \\ 
After unit and hub checks & 48{,}600 & 26.4\% & 7.1\% \\ 
After leakage check & 41{,}200 & 25.8\% & 5.4\% \\ 
After novelty check (candidates) & 38{,}900 & 25.5\% & 0.0\% \\ 
\quad $\hat p=1$, discarded & 8{,}558 & 14.2\% & -- \\ 
\quad $0<\hat p<1$, RL route & 17{,}116 & 26.0\% & -- \\ 
\quad $\hat p=0$, demonstration route & 13{,}226 & 31.9\% & -- \\ 
\rowcolor{PastaYellow}
Admitted for RL & 12{,}400 & 27.1\% & 0.0\% \\ 
\rowcolor{PastaYellow}
Self demonstrated & 4{,}007 & 38.6\% & 0.0\% \\ 
\bottomrule
\end{tabular}}
\vspace{-2mm}
\end{table}

\begin{table}[!t]
\centering
\caption{Cost of world expansion (9B, full run on 16 NVIDIA H20 GPUs).}
\label{tab:cost}
\vspace{-2mm}
\resizebox{0.95\linewidth}{!}{
\begin{tabular}{l|cc}
\toprule
\textbf{Cost Item} & \textbf{Static Growth} & \textbf{Search-RSI} \\
\midrule
Training rollouts & 102.4k & 102.4k \\ 
Probe rollouts (truncated) & 0 & 155.6k \\ 
Hinted retries (truncated) & 0 & 52.9k \\ 
Supervised tokens per round & 0 & \hyp{app.cost.tok}{6.1M} \\ 
Injected facts & 0 & 27.2k \\ 
Wall clock per round (h) & 5.9 & 7.6 \\ 
Peak index storage (GB) & 41.8 & 42.1 \\ 
Retrieval calls (M) & 1.84 & 2.71 \\ 
Relative GPU hours & 1.00$\times$ & 1.27$\times$ \\ 
\bottomrule
\end{tabular}}
\vspace{-2mm}
\end{table}

\input{appendix_robust}

\section{Limitations}
\textbf{Scope of the claim.} Our evidence shows that, in the growth protocol we study, the static training world is a main bottleneck of online task growth and that expanding it along the frontier removes it. We do not claim that this is the only bottleneck of multimodal search agents, and the analyses of Appendix~\ref{app:robust} indicate where the gain is small, for example on single entity recognition and on events after the graph snapshot.

\textbf{Bounded and pipeline specific world.} \ours expands the world only within the coverage of the source graph, so the recursion is bounded: once the frontier units exhaust the paths that $\mathcal{G}$ can support, further rounds add little. All our runs use one graph; the shifts of renderer, verbalization, retriever, and relation family in Table~\ref{tab:pipeline_shift} reduce but do not remove this dependence, and a second source graph is the most important open test. The renderer is frozen by design, which keeps the attribution of gains clean but also caps the linguistic diversity of new questions. The requirements are concrete: a source from which a supporting path can be sampled, a resolver that turns each element of that path into an indexed document, and write access to the training index. Recipes that already synthesize tasks from a graph meet all three, so the loop transfers to them as Table~\ref{tab:transfer} shows. Settings that retrieve from the open web meet none of them, and Appendix~\ref{app:robust} shows that unreachable negatives persist there, so a local evidence cache beside the web index, rather than the method as it stands, is what such settings would need.

\textbf{Recursion.} Our evidence supports a loop in which the improver tracks the policy being trained, not an open ended recursion that compounds without limit. Replacing the improver by a stronger external agent changes the result by at most \hyp{lim.teacher}{0.1}, and the per round gain in Fig.~\ref{fig:scaling_res} is sublinear over ten rounds, so we describe \ours as policy in the loop world expansion and leave longer horizons, where the graph itself becomes the binding constraint, to future work.

\textbf{Privileged hint.} Self demonstration is hint guided rather than autonomous: the hint is built from the path of the task and narrows the search, as Table~\ref{tab:hint} quantifies. The anchor first check guards against naming hidden entities before retrieving them but cannot rule out subtler forms of guidance, which is why the hint never enters training and the RL phase remains the main source of updates.

\textbf{Design heuristics.} The mastery band, the $3{\times}3{\times}3$ discretization, the hop disagreement score, $K$, and $M$ are set empirically. The sensitivity analyses show broad optima, but we present \ours as a training recipe rather than a general learning principle.

\input{appendix_cases}

%% file: tables/task_unit_description.tex
\begin{tabular}{l|l|l|l|l}
\toprule
\textbf{Axis} & \textbf{Definition} & \textbf{Example (low)} & \textbf{Example (high)} & \textbf{Typical Failure} \\
\midrule
Hop depth $h$ & Number of relations on the supporting path $\pi_x$ & \textit{stadium} $\rightarrow$ \textit{architect} ($h{=}2$) & \textit{stadium} $\rightarrow$ \textit{architect} $\rightarrow$ \textit{firm} $\rightarrow$ \textit{founding year} ($h{=}4$) & Evidence chain breaks at a late hop \\
Relation rarity $r$ & Mean negative log frequency of the relations on $\pi_x$ & \textit{located in}, \textit{born in} & \textit{structural engineer of}, \textit{successor firm of} & Query for the relation is malformed \\
Distractor density $d$ & Largest set of same type entities near a path node & A landmark with a unique appearance & One of many look alike suspension bridges & Agent commits to a look alike entity \\
\midrule
Unit $c=(h,r,d)$ & One cell of the discretized space $\mathcal{U}$ ($3{\times}3{\times}3=27$ units) & \multicolumn{2}{l|}{Tasks in one unit share a failure cause, so unit level statistics localize \textit{why} the agent fails} & Unsupported path: \textit{unreachable negative} \\
\bottomrule
\end{tabular}

%% file: appendix_robust.tex
\section{Robustness and Attribution Analyses}
\label{app:robust}
This section examines the questions that the main results leave open: whether the gains are stable across seeds, whether a simpler form of world repair would suffice, whether the gain comes from the world or from extra training signal, whether support injection makes retrieval easier rather than more consistent, whether the gains survive outside the pipeline that builds the training world, how much of self demonstration is owed to the privileged hint, whether the method transfers to another recipe and backbone, what it costs at training and test time, how far environment consistency can be trusted as a metric, and whether the problem persists on the open web. It closes with an analysis of the failures that remain.

\subsection{Seed Variance and Significance}
We repeat the full method, static world growth, and the four ablations that carry our mechanism claims with three seeds at 9B. The seed changes path sampling, probe rollouts, hinted retries, and RL sampling, which are the stochastic parts of the loop. Table~\ref{tab:seeds} reports means and standard deviations, and, for every variant, the paired bootstrap 95\% confidence interval of its gap to \ours on the six benchmark average (10{,}000 resamples, clustered by seed so that the three runs of a variant are resampled as units rather than pooled, which would treble the apparent sample size and understate the variance) with Holm corrected $p$ values over the five comparisons. 

The standard deviation of the average stays at or below \hyp{app.seed.std}{0.5} for every variant, an order of magnitude below the gap between \ours and static world growth, and all five gaps exclude zero after correction. The beyond frontier bucket is the noisiest column, as expected for the longest and rarest tasks, yet its gap to the variant without self demonstration remains \hyp{app.seed.sdb}{5.8} points. At 27B we run three seeds for the full method and for static world growth, with standard deviations of \hyp{app.seed.std27}{0.4} and \hyp{app.seed.std27s}{0.5} on the average, and a single seed for every ablation and for the analyses of this section, which are therefore reported at 9B only. All answers are judged by the same GPT-5 prompt, and on \hyp{app.seed.nhuman}{300} answers sampled across benchmarks the judge agrees with two human annotators on \hyp{app.seed.agree}{97.3}\% of them, so judge noise is small relative to the reported gaps. 

\textbf{Evaluation protocol.} Two scoring rules are used and never mixed within a table. All public benchmarks in Tables~\ref{tab:main_res} and \ref{tab:main_full} are scored by the GPT-5 answer judge \cite{singh2025openai} with the fixed prompt of Appendix~B, against the released answer sets. All in world diagnostics (Table~\ref{tab:diagnostic_res}, the independent split, and the frontier buckets) are scored by exact match against the alias set of the answer entity, which needs no judge. To check that the judge does not favor any method, we re-score the controlled block of Table~\ref{tab:main_res} with a second judge (Gemini-3.1-Pro \cite{google2026gemini31}) using the same prompt: the ranking of all controlled methods is unchanged, and the gap between \ours and static world growth moves from \hyp{app.judge.g1}{10.4} to \hyp{app.judge.g2}{10.1} points at 9B. 

\begin{table}[!t]
\centering
\caption{Seed variance (9B, three seeds, mean $\pm$ std). The last column is the gap to \ours on Avg.\ with its paired bootstrap 95\% interval; $^{*}$ marks Holm corrected $p<0.01$.}
\label{tab:seeds}
\vspace{-2mm}
\resizebox{\linewidth}{!}{
\begin{tabular}{l|ccc|c|c}
\toprule
\textbf{Variant} & \textbf{Avg.} & \textbf{Indep.} & \textbf{Beyond} & \textbf{Cons} & \textbf{Gap [95\% CI]} \\
\midrule
Static World Growth & 58.7\std{0.5} & 69.2\std{0.6} & 38.7\std{0.9} & 0.85\std{.01} & 10.4 [9.6, 11.2]$^{*}$ \\ 
\textit{w/o} Support Injection & 65.5\std{0.4} & 76.0\std{0.5} & 48.9\std{0.8} & 0.93\std{.01} & 3.6 [2.9, 4.3]$^{*}$ \\ 
\textit{w/o} Frontier Diagnosis & 66.5\std{0.4} & 77.9\std{0.4} & 51.8\std{0.7} & 0.98\std{.00} & 2.6 [1.9, 3.3]$^{*}$ \\ 
\textit{w/o} Self Demonstration & 67.1\std{0.3} & 78.3\std{0.4} & 48.8\std{0.6} & 0.98\std{.00} & 2.0 [1.4, 2.6]$^{*}$ \\ 
Frozen Improver ($\pi_0$) & 65.9\std{0.4} & 76.6\std{0.5} & 49.7\std{0.8} & 0.98\std{.00} & 3.2 [2.5, 3.9]$^{*}$ \\ 
\rowcolor{PastaYellow}
\textbf{Search-RSI (Ours)} & \textbf{69.1}\std{0.3} & \textbf{79.6}\std{0.4} & \textbf{54.6}\std{0.6} & 0.98\std{.00} & -- \\ 
\bottomrule
\end{tabular}}
\vspace{-3mm}
\end{table}

\subsection{Stronger World Expansion Baselines}
Static world growth is the natural baseline because it is what current recipes do, but it is also the setting in which unreachable negatives arise by construction. We therefore compare with four baselines that also repair or expand the world, each removing a different part of the recursive loop (Table~\ref{tab:strong_baselines}). \textit{Support constrained generation} prevents the problem instead of repairing it, by adding $\Psi(\gamma)\subseteq\knowledge_t$ as a fifth acceptance condition in Eq.~(\ref{eq:accept}), so no injection is needed and the world is consistent by construction. \textit{Retriever only repair} keeps the static task growth and, after each round, patches the index with the missing facts of the queries that failed, so repair happens on the evidence side only. \textit{Random support repair} injects the support of every admitted task but uses uniform quotas and random retention without self demonstration, so the world is consistent but the growth is not directed. \textit{Untargeted support} injects as many facts as \ours does, drawn from random paths of $\mathcal{G}$ rather than from the admitted tasks. \textit{Teacher guided admission} replaces the probes and routing of the current policy by those of an external 27B agent trained with the same recipe. The frozen improver and the oracle complete index from Tables~\ref{tab:ablation_study} and \ref{tab:control} complete the comparison. 

Four findings follow. First, prevention is not free. Support constrained generation reaches a consistency of one without any injection, but Table~\ref{tab:audit} shows why it is not enough: after the four acceptance conditions of Eq.~(\ref{eq:accept}) the share of candidates with missing support barely moves, from \hyp{app.sb.f0}{29.0}\% to \hyp{app.sb.f1}{25.5}\%, because missing support concentrates in the deep and rare paths that the frontier consists of. Requiring indexed support therefore removes those paths: the share of admitted tasks in frontier units falls to \hyp{app.sb.scgfront}{24.8}\% and the average stops at \hyp{app.sb.scg}{60.9}, below every protocol that injects support. Among the admitted tasks of \ours, \hyp{app.sb.dep1}{27.1}\% rely on injected support, and among the self demonstrated ones \hyp{app.sb.dep2}{38.6}\%. Second, consistency is necessary but not sufficient: retriever only and random support repair raise the consistency to \hyp{app.sb.cons1}{0.93} and \hyp{app.sb.cons2}{0.98} and recover most of the gap on the independent split, but they trail \ours by \hyp{app.sb.g1}{8.1} and \hyp{app.sb.g2}{7.0} points on the average and by more on the beyond frontier bucket, because neither decides where the world should grow. Third, the volume of added evidence is not what matters: untargeted support adds the same number of facts and leaves both consistency and accuracy almost unchanged, since the missing facts are specific to the new tasks. Fourth, letting an external agent route the candidates is slightly worse than letting the policy route them itself (\hyp{app.sb.teach}{68.6} vs.\ 69.1), which supports the premise of the recursion: the frontier that matters is the frontier of the policy being trained. 

\begin{table}[!t]
\centering
\caption{World expansion baselines (9B). All variants share the backbone, optimizer, tools, and task budget. Front.\ is the share of admitted tasks in frontier units; Beyond and Indep.\ are exact match in the frozen index $\Keval$.}
\label{tab:strong_baselines}
\vspace{-2mm}
\resizebox{\linewidth}{!}{
\begin{tabular}{l|cc|cccc}
\toprule
\textbf{Protocol} & \textbf{Cons} & \textbf{UN (\%)} & \textbf{Front.} & \textbf{Beyond} & \textbf{Indep.} & \textbf{Avg.} \\
\midrule
Static World Growth & 0.85 & 41.0 & 31.5 & 38.7 & 69.2 & 58.7 \\ 
Untargeted support (matched volume) & 0.86 & 37.9 & 31.9 & 39.2 & 69.6 & 59.1 \\ 
Support constrained generation & 1.00 & 0.0 & 24.8 & 40.1 & 70.8 & 60.9 \\ 
Retriever only repair & 0.93 & 14.6 & 32.4 & 41.9 & 71.4 & 61.0 \\ 
Random support repair & 0.98 & 4.6 & 33.9 & 43.0 & 72.8 & 62.1 \\ 
Non recursive expansion (frozen $\pi_0$) & 0.98 & 4.6 & 37.5 & 49.7 & 76.6 & 65.9 \\ 
Oracle complete index & 1.00 & 0.0 & 34.6 & 46.1 & 74.0 & 63.9 \\ 
Teacher guided admission & 0.98 & 4.4 & 44.0 & 54.2 & 79.2 & 68.6 \\ 
\rowcolor{PastaYellow}
\textbf{Search-RSI (Ours)} & 0.98 & 4.4 & \textbf{44.2} & \textbf{54.6} & \textbf{79.6} & \textbf{69.1} \\ 
\bottomrule
\end{tabular}}
\vspace{-3mm}
\end{table}

\subsection{World Expansion versus Extra Training Signal}
Compared with static world growth, \ours changes the world but also adds training signal: \hyp{app.sig.demos}{4.0}k self demonstrations, one supervised step per round, and probing and hinted retries that cost $1.27\times$ GPU hours in total. Table~\ref{tab:signal} separates these sources by matching each of them on top of static world growth, and by removing all of them from \ours. Matching the compute alone, matching the supervised data with \hyp{app.sig.sft}{4.0}k replayed warm up trajectories, or adding self demonstrations to the static world each recovers at most \hyp{app.sig.max}{1.4} of the \hyp{app.sig.gap}{10.4} point gap. Conversely, \ours without self demonstration, which receives no supervised data beyond the warm up and runs at the compute of static growth plus probing, still gains \hyp{app.sig.world}{8.4} points; giving that variant the remaining compute adds only \hyp{app.sig.wc}{0.1}. Most of the gain therefore comes from the expanded world, and the extra supervised signal contributes the remaining \hyp{app.sig.sd}{2.0} points, concentrated on the beyond frontier bucket. 

\begin{table}[!t]
\centering
\caption{Separating the world from the extra training signal (9B). SFT data: supervised trajectories added on top of the warm up; Compute: GPU hours relative to static world growth.}
\label{tab:signal}
\vspace{-2mm}
\resizebox{\linewidth}{!}{
\begin{tabular}{l|cc|cc}
\toprule
\textbf{Variant} & \textbf{SFT data} & \textbf{Compute} & \textbf{Beyond} & \textbf{Avg.} \\
\midrule
Static World Growth & 0 & 1.00$\times$ & 38.7 & 58.7 \\ 
\quad + matched compute ($1.27\times$ steps) & 0 & 1.27$\times$ & 38.9 & 58.8 \\ 
\quad + matched SFT data (replayed warm up) & 4.0k & 1.27$\times$ & 39.6 & 59.0 \\ 
\quad + self demonstrations in the static world & 4.0k & 1.27$\times$ & 42.0 & 60.1 \\ 
\midrule
\ours \textit{w/o} Self Demonstration & 0 & 1.19$\times$ & 48.8 & 67.1 \\ 
\quad + remaining compute (more RL steps) & 0 & 1.27$\times$ & 49.0 & 67.2 \\ 
\rowcolor{PastaYellow}
\textbf{\ours (Full)} & 4.0k & 1.27$\times$ & \textbf{54.6} & \textbf{69.1} \\ 
\bottomrule
\end{tabular}}
\vspace{-3mm}
\end{table}

\textbf{Probe rollouts as a budget.} Probing spends \hyp{app.sig.probe}{155.6}k truncated rollouts, more than the \hyp{app.sig.train}{102.4}k rollouts used for policy updates. These rollouts never enter the training buffer, but they are on policy samples, so a fair question is what static world growth would do with the same sampling budget. Two matched variants answer it: raising its group size from 8 to 12, which spends a comparable number of rollouts per step, reaches \hyp{app.sig.grp}{59.0}, and growing \hyp{app.sig.tasks}{1.5}$\times$ as many tasks per round at the same number of steps reaches \hyp{app.sig.more}{58.4}, slightly below the default because the added tasks also carry unreachable negatives. Extra sampling in a static world therefore does not substitute for a consistent one. 

\begin{figure*}[!t]
    \centering
    \includegraphics[width=0.96\linewidth]{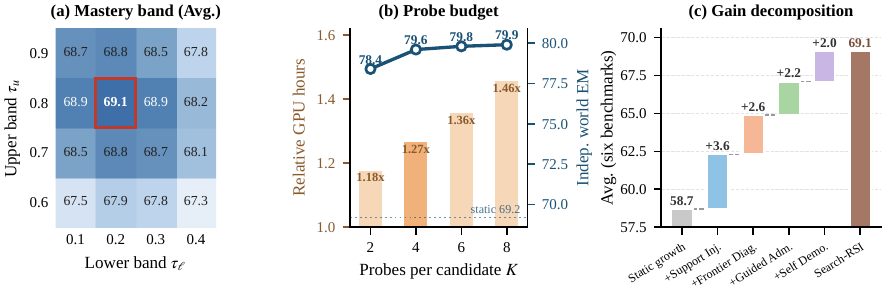}
    \vspace{-3mm}
    \caption{Sensitivity analysis (9B). (a) Six benchmark average over the mastery band $[\tau_\ell,\tau_u]$; the red box marks the default $[0.2,0.8]$. (b) Probe budget $K$: bars give relative GPU hours and the line gives EM on the independent world split. (c) Cumulative gain over static world growth as Support Injection, Frontier Diagnosis, Guided Admission and Self Demonstration are added in turn.}
    \vspace{-3mm}
    \label{fig:sensitivity}
\end{figure*}

\subsection{Does Support Injection Simplify Retrieval?}
Because the injected facts are tied to the admitted tasks, one may suspect that \ours tailors the training world to its own tasks and makes retrieval easier rather than merely consistent. Table~\ref{tab:retrieval_difficulty} measures the difficulty of the training index directly. After ten rounds, \ours has added \hyp{app.rd.facts}{27.2}k facts, about \hyp{app.rd.share}{0.6}\% of the index, so the bulk of the evidence and of its distractors is unchanged. The number of look alike entities that a query must disambiguate (entities of the same type that share a name token or a visual cluster with the target) rises slightly instead of falling, because new support also enlarges the neighborhoods of existing entities. The rate at which the first query of a task retrieves its gold evidence in the top five results on newly admitted tasks matches that of the round 0 pool, while static growth lowers it because many gold facts are missing. Finally, the number of tool calls per task increases. Injection therefore restores answerability without making the admitted tasks easier to retrieve than the original pool, and all reported accuracies are measured in the frozen evaluation index, which never receives injected facts. 

\begin{table}[!t]
\centering
\caption{Difficulty of the training index after ten rounds (9B). Look alike: same type distractors per query; Gold@5: first query retrieves gold evidence in the top five, on the tasks added in the last round (round 0: initial pool).}
\label{tab:retrieval_difficulty}
\vspace{-2mm}
\resizebox{\linewidth}{!}{
\begin{tabular}{l|cccc}
\toprule
\textbf{Index} & \textbf{Facts (M)} & \textbf{Look alike} & \textbf{Gold@5 (\%)} & \textbf{Calls / task} \\
\midrule
Round 0 index & 4.21 & 6.3 & 41.2 & 3.4 \\ 
Static World Growth & 4.21 & 6.3 & 33.5 & 3.1 \\ 
Commit all staged support & 4.36 & 7.4 & 39.8 & 4.3 \\ 
Oracle complete index & 5.02 & 9.1 & 38.1 & 4.4 \\ 
\rowcolor{PastaYellow}
\textbf{Search-RSI (Ours)} & 4.24 & 6.5 & 40.6 & 4.1 \\ 
\bottomrule
\end{tabular}}
\vspace{-3mm}
\end{table}

\subsection{Shifts of the Construction Pipeline}
The training world is grown by one graph, one renderer, one verbalization of support statements, and one retriever. To test whether the agent has only adapted to this pipeline, we change each of these components at evaluation time while keeping the trained checkpoints fixed (Table~\ref{tab:pipeline_shift}). \textit{Held out relations} restricts the independent split to questions whose paths use only relation families that were never sampled during growth. \textit{New renderer} rewrites the evaluation questions with a different model and prompt. \textit{Paraphrased support} rewrites every relation statement of $\Keval$ with a different verbalization style. \textit{BM25 retriever} replaces the dense retriever of the text search tool by a sparse one. The public benchmarks of Table~\ref{tab:main_res} use their own backends and are listed for reference. 

The gain shrinks under every shift but stays between \hyp{app.ps.min}{7.3} and \hyp{app.ps.max}{9.3} points. It is smallest on held out relations, where the frontier diagnosis never targeted the relevant units, and largest under a new renderer, which suggests that the agent did not learn the surface form of the questions. We regard a second source graph as the most important remaining test, which we discuss in the limitations. 

\begin{table}[!t]
\centering
\caption{Gains under shifts of the construction pipeline (9B, EM). Checkpoints are unchanged; only the evaluation side is altered.}
\label{tab:pipeline_shift}
\vspace{-2mm}
\resizebox{0.92\linewidth}{!}{
\begin{tabular}{l|cc|c}
\toprule
\textbf{Evaluation Setting} & \textbf{Static} & \textbf{Search-RSI} & $\Delta$ \\
\midrule
Independent split (default) & 69.2 & 79.6 & \up{10.4} \\ 
Held out relation families & 61.4 & 68.7 & \up{7.3} \\ 
New renderer & 68.4 & 77.7 & \up{9.3} \\ 
Paraphrased support & 67.8 & 76.6 & \up{8.8} \\ 
BM25 retriever & 66.0 & 75.1 & \up{9.1} \\ 
\midrule
Public benchmarks (Avg., Table~\ref{tab:main_res}) & 58.7 & 69.1 & \up{10.4} \\ 
\bottomrule
\end{tabular}}
\vspace{-3mm}
\end{table}

\subsection{Role of the Privileged Hint}
Self demonstration is not unassisted: the retry sees a hint $\eta(x)$ derived from the path of the task, which narrows the search space even though it never contains the answer. Table~\ref{tab:hint} varies the content of the hint. Without any hint, plain retries rarely succeed on tasks that every probe failed, and the few kept demonstrations add little over the variant without self demonstration. Giving only the relation chain recovers about half of the gain, giving only the bridge entities recovers more, and the full chain is best. Demonstrations from an external 27B agent are only \hyp{app.hint.teach}{0.1} better than those of the policy itself. Three sanity checks probe whether the hint acts through hidden guidance rather than through the retrieved evidence. A wrong hint, taken from the chain of another task, and a random relation chain both yield almost no kept demonstrations and no gain, so the benefit requires a hint that matches the task. Revealing only the number of hops is no better than plain retries, and shuffling the order of a correct chain keeps only a small part of the gain, so the hint helps by fixing the order of hops to resolve, while the content of each hop must still be retrieved. Removing the anchor first check raises the yield to \hyp{app.hint.noaf}{58}\% but lowers the average to \hyp{app.hint.noafavg}{67.5}: the extra trajectories name hidden entities before retrieving them, and imitating them teaches the policy to guess. The check is therefore what turns hinted retries into useful demonstrations. We therefore describe self demonstration as hint guided self distillation: it uses no stronger model, but it does rely on privileged information about the task, which is removed before training and is never available at test time. 

\begin{table}[!t]
\centering
\caption{Content of the privileged hint (9B). Yield is the share of $\hat p=0$ candidates turned into kept demonstrations in the last round.}
\label{tab:hint}
\vspace{-2mm}
\resizebox{0.95\linewidth}{!}{
\begin{tabular}{l|cc|cc}
\toprule
\textbf{Hint} & \textbf{Yield (\%)} & \textbf{Demos} & \textbf{Beyond} & \textbf{Avg.} \\
\midrule
None (\textit{w/o} Self Demonstration) & -- & 0 & 48.8 & 67.1 \\ 
No hint, plain retries & 9 & 0.9k & 49.6 & 67.2 \\ 
Relation chain only & 27 & 2.6k & 52.1 & 68.3 \\ 
Bridge entities only & 33 & 3.1k & 53.0 & 68.7 \\ 
\rowcolor{PastaYellow}
\textbf{Full chain (Ours)} & 41 & 4.0k & 54.6 & 69.1 \\ 
External teacher demonstrations & -- & 4.0k & 54.9 & 69.2 \\ 
\midrule
\multicolumn{5}{l}{\textit{Sanity checks}} \\
Wrong hint (chain of another task) & 3 & 0.3k & 48.9 & 67.1 \\ 
Random relation chain & 5 & 0.5k & 49.2 & 67.1 \\ 
Hop count only & 11 & 1.1k & 49.7 & 67.2 \\ 
Correct chain, shuffled order & 14 & 1.3k & 50.2 & 67.4 \\ 
Full chain, anchor first check removed & 58 & 5.6k & 50.8 & 67.5 \\ 
\bottomrule
\end{tabular}}
\vspace{-3mm}
\end{table}

\subsection{Transfer across Recipes and Backbones}
All main experiments build on the \textsc{SearchEyes} recipe. To test whether \ours depends on it, we apply the same loop, with the same hyperparameters, to two further settings at a reduced budget of \hyp{app.tr.rounds}{5} rounds (Table~\ref{tab:transfer}). The first replaces the recipe: we start from the warm up checkpoint and use the graph, synthesis pipeline, and tool interface of OpenSearch-VL \cite{chen2026opensearch} with the Qwen3.5-9B backbone. The second replaces the backbone family with Qwen3-VL-8B \cite{bai2025qwen3} under the \textsc{SearchEyes} recipe. In both settings static world growth again loses consistency over rounds, and \ours again recovers it and improves the six benchmark average, by \hyp{app.tr.g1}{6.9} and \hyp{app.tr.g2}{7.5} points. The gains are smaller than in the main setting, partly because of the shorter run, but they show that the effect is not specific to one recipe or one backbone. 

\begin{table}[!t]
\centering
\caption{Transfer to another recipe and another backbone (5 rounds, Avg.\ over the six standard benchmarks).}
\label{tab:transfer}
\vspace{-2mm}
\resizebox{\linewidth}{!}{
\begin{tabular}{l|cc|cc|c}
\toprule
& \multicolumn{2}{c|}{\textbf{Static Growth}} & \multicolumn{2}{c|}{\textbf{Search-RSI}} & \\
\textbf{Setting} & \textbf{Cons} & \textbf{Avg.} & \textbf{Cons} & \textbf{Avg.} & $\Delta$ \\
\midrule
\textsc{SearchEyes}, Qwen3.5-9B (main, 5 rounds) & 0.89 & 60.2 & 0.97 & 67.5 & \up{7.3} \\ 
OpenSearch-VL recipe, Qwen3.5-9B & 0.88 & 57.1 & 0.97 & 64.0 & \up{6.9} \\ 
\textsc{SearchEyes}, Qwen3-VL-8B & 0.89 & 55.4 & 0.97 & 62.9 & \up{7.5} \\ 
\bottomrule
\end{tabular}}
\vspace{-3mm}
\end{table}

\subsection{Cost and Inference Efficiency}
\ours costs more to train and its agents search longer at test time. Table~\ref{tab:cost_benefit} relates both costs to the gain. Training takes $1.27\times$ the GPU hours of static world growth on the same 16 H20 GPUs, so the gain per additional GPU hour is large compared with spending the same compute on more RL steps, which adds \hyp{app.cb.cm}{0.1} (Table~\ref{tab:signal}). At test time, \ours issues \hyp{app.cb.calls}{1.3} more tool calls per question on average and its median latency rises from \hyp{app.cb.l0}{38} to \hyp{app.cb.l1}{47} seconds. To separate the gain from the extra search, we cap the number of tool calls at the average budget of static growth: \ours still reaches \hyp{app.cb.cap}{66.0}, so about seven tenths of its gain does not require longer search. A smaller probe budget, $K{=}2$, lowers the training overhead to $1.18\times$ at a cost of \hyp{app.cb.k2}{1.7} points, which gives a cheaper operating point on the same Pareto front as Fig.~\ref{fig:sensitivity}(b). 

\begin{table}[!t]
\centering
\caption{Cost and benefit (9B). Training cost relative to static world growth; inference measured on the six standard benchmarks.}
\label{tab:cost_benefit}
\vspace{-2mm}
\resizebox{\linewidth}{!}{
\begin{tabular}{l|c|cc|c}
\toprule
\textbf{Variant} & \textbf{Train cost} & \textbf{Calls / q} & \textbf{Latency (s)} & \textbf{Avg.} \\
\midrule
Static World Growth & 1.00$\times$ & 4.0 & 38 & 58.7 \\ 
Static World Growth ($1.27\times$ steps) & 1.27$\times$ & 4.0 & 38 & 58.8 \\ 
\ours, $K{=}2$ & 1.18$\times$ & 5.1 & 45 & 67.4 \\ 
\ours, calls capped at 4.0 & 1.27$\times$ & 4.0 & 37 & 66.0 \\ 
\rowcolor{PastaYellow}
\textbf{\ours} ($K{=}4$) & 1.27$\times$ & 5.3 & 47 & \textbf{69.1} \\ 
\bottomrule
\end{tabular}}
\vspace{-3mm}
\end{table}

\subsection{Environment Consistency as a Metric}
Environment consistency, $\mathrm{Cons}_t$ in Eq.~(\ref{eq:consistency}), is the fraction of tasks in the pool whose support set is fully retrievable from the training index, where an element counts as retrievable if a query formed from its entity name and relation returns it in the top ten results. To check this automatic test, we manually verified \hyp{app.cons.n}{500} tasks sampled across rounds and methods; the automatic label agrees with the manual one on \hyp{app.cons.agree}{96.2}\% of them, and almost all disagreements are support statements that are present but phrased so that the canonical query misses them. Across the fifteen protocols of Tables~\ref{tab:ablation_study}, \ref{tab:control}, and \ref{tab:strong_baselines}, $\mathrm{Cons}$ and the six benchmark average have a Pearson correlation of \hyp{app.cons.r}{0.86}. The relation is one sided, however: every protocol with low consistency is weak, but protocols with consistency of at least 0.98 range from \hyp{app.cons.lo}{62.1} to \hyp{app.cons.hi}{69.2}. We therefore use consistency as a necessary condition and a diagnostic, not as evidence of search ability by itself; the latter is measured by accuracy in the frozen evaluation index and on the public benchmarks. 

\subsection{Beyond Graph Bounded Worlds}
World expansion as implemented here relies on a source graph, a renderer, and a resolver that maps support statements to documents, so it applies most directly to synthesized training worlds. To estimate whether unreachable negatives also matter when training retrieves from the open web, we replace the training index by a live web search tool for the first three rounds of static world growth. Even then, \hyp{app.web.un}{18.4}\% of newly grown tasks have at least one support element that the top ten web results never contain, typically a relation between two rare entities stated only on a page that the search engine ranks low, and these tasks account for \hyp{app.web.neg}{26.7}\% of negative rewards. The problem is thus smaller on the open web but not absent. Since support cannot be injected into a web index, a practical variant would route such tasks to a local evidence cache; we leave this to future work and restrict our claims to graph grounded training worlds. 

\subsection{Failure Analysis}
\textbf{Misrouting.} To estimate how often the probes route a candidate wrongly, we re-probe \hyp{app.fa.n}{2{,}000} candidates with $K{=}32$ rollouts and treat the resulting success rate $p^{\star}$ as the reference. Of the candidates discarded as mastered, \hyp{app.fa.p1}{4.8}\% have $p^{\star}<0.5$; of those routed to RL, \hyp{app.fa.rl1}{7.9}\% have $p^{\star}\ge0.95$ and \hyp{app.fa.rl0}{3.1}\% have $p^{\star}\le0.03$, so they contribute almost no gradient; and of those routed to self demonstration, \hyp{app.fa.sd}{12.6}\% have $p^{\star}\ge0.25$ and could have been learned by RL. The last error is the most common, which is the expected cost of $K{=}4$ and is consistent with the sensitivity of the probe budget in Fig.~\ref{fig:sensitivity}. 

\textbf{Where the gain is smallest.} Per benchmark, the 9B gain over static world growth is smallest on \textsc{SimpleVQA} (\hyp{app.fa.svqa}{+7.0}), InfoSeek (\hyp{app.fa.info}{+8.1}), and \textsc{LiveVQA} (\hyp{app.fa.live}{+8.7}), and largest on BC-V3 (\hyp{app.fa.bc}{+14.0}), BC-VL (\hyp{app.fa.bcvl}{+13.7}), and \textsc{VDR} (\hyp{app.fa.vdr}{+13.6}). The first two are dominated by recognizing a single entity, which world expansion does not address, and \textsc{LiveVQA} asks about events after the snapshot of $\mathcal{G}$. The gain is largest where questions chain several hops through rare relations, which is where unreachable negatives concentrate. 

\textbf{What stays unreachable.} Injection repairs the support of a candidate only when the resolver finds a document for every element of its path. For \hyp{app.fa.unrep}{6.3}\% of the candidates it does not: the path passes through a relation that $\mathcal{G}$ states as a bare triple with no page behind it, or through an entity whose page exists but names the relation only in an image caption that the text index does not hold. These candidates stay in the staging area and are never admitted, so their unreachability is avoided rather than removed. By unit, they concentrate in the rarest relation tier at $h{=}4$, where \hyp{app.fa.unrep.unit}{19.4}\% of candidates cannot be repaired, which is also where the frontier diagnosis would most like to grow. A second failure mode survives repair: \hyp{app.fa.hardunit}{11.2}\% of the frontier units receive their full quota for ten rounds and end the run below the mastery band, almost all of them with four hops and a high distractor density. The quota rule keeps feeding them because their rollouts still disagree, so a unit level stopping rule is the clearest extension we did not implement.

\textbf{Remaining errors.} We manually label \hyp{app.fa.nlab}{300} failures of each agent on the independent split (Table~\ref{tab:errors}). \ours mostly removes early stopping, the error that unreachable negatives teach, so recognition errors and confusion with look alike entities become the dominant failures. Evidence that is absent even from $\Keval$ accounts for a similar share in both agents, which confirms that world expansion during training does not change the evaluation world. 

\textbf{When expansion adds noise.} Expansion is not free of side effects. Committing support for every staged candidate lowers the average (Table~\ref{tab:control}) because unused facts enlarge the neighborhoods of look alike entities, and even selective commitment raises the look alike count slightly over ten rounds (Table~\ref{tab:retrieval_difficulty}). In addition, \hyp{app.fa.conf}{0.7}\% of the injected facts conflict with an existing statement for the same entity and relation, typically a different date or an outdated affiliation; tasks whose support contains such a conflict have a \hyp{app.fa.confgap}{9.2} point lower success rate. We do not filter them in this work. Since these tasks are \hyp{app.fa.confshare}{0.9}\% of the admitted pool and lag by \hyp{app.fa.confgap}{9.2} points, removing them would change the average by less than \hyp{app.fa.confeff}{0.1}, so a consistency check against the existing index is worthwhile for correctness rather than for accuracy. 

\begin{table}[!t]
\centering
\caption{Error types of 300 sampled failures per agent on the independent split (9B, \%).}
\label{tab:errors}
\vspace{-2mm}
\resizebox{0.86\linewidth}{!}{
\begin{tabular}{l|cc}
\toprule
\textbf{Error Type} & \textbf{Static Growth} & \textbf{Search-RSI} \\
\midrule
Early stop or answer from memory & 27 & 12 \\ 
Visual recognition of the anchor & 24 & 31 \\ 
Look alike entity confusion & 17 & 22 \\ 
Wrong relation at an intermediate hop & 13 & 15 \\ 
Evidence absent from $\Keval$ & 8 & 9 \\ 
Answer format or judge error & 6 & 6 \\ 
Other & 5 & 5 \\ 
\bottomrule
\end{tabular}}
\vspace{-3mm}
\end{table}

%% file: appendix_cases.tex
\section{Case Studies}
\label{app:cases}
This section walks through four questions from VisBrowse \cite{zhang2026visbrowse}, each with the image released in the official benchmark repository. We chose items whose images show no identifiable person and whose reasoning chains exercise different parts of \ours. For every case we give the benchmark item, the image, the question and the ground truth exactly as released, and then compare the trajectory of the agent trained by static world growth with that of the 9B agent trained by \ours. Trajectories are abbreviated: we keep the reasoning that decides the next action, every tool call, and the part of each observation that the agent uses. 

The failures of the static world baseline fall into three recurring patterns. The first is \textit{early answering}: once the agent believes it knows an entity, it answers a fine grained attribute from memory instead of retrieving it. The second is \textit{coarse grounding}: the agent searches the whole image when the question points to one region, and recovers the wrong entity. The third is \textit{constraint dropping}: in a chain of three or more conditions, one condition is silently ignored, so the answer satisfies a simpler question. All three are consistent with the error taxonomy of Appendix~\ref{app:robust} and with RQ3 in the main paper: a policy that is repeatedly penalized when its evidence is missing learns that further searching does not pay off. \ours keeps the reward of a correct search intact through support injection, admits tasks whose difficulty lies in a particular hop, and turns tasks beyond reach into demonstrations that resolve every hop with retrieved evidence.

\newcommand{\caseq}[1]{\textcolor{deltared}{\textbf{#1}}}
\newcommand{\tool}[1]{\texttt{\textcolor{cvprblue}{#1}}}
\newcommand{\okmark}{\textcolor{darkgreen}{\ding{51}}}
\newcommand{\nomark}{\textcolor{deltared}{\ding{55}}}

\subsection{Case 1: Retrieve, Do Not Recall}
The first case needs only two hops, but the final attribute is a visual detail of a logo that is easy to misremember.

\begin{tcolorbox}[title=Case 1: VisBrowse item 117 (Technology),width=\linewidth,breakable,colback=white,colframe=black!55,fonttitle=\small\bfseries,left=3pt,right=3pt]
\begin{small}
\begin{center}
\includegraphics[width=0.86\linewidth]{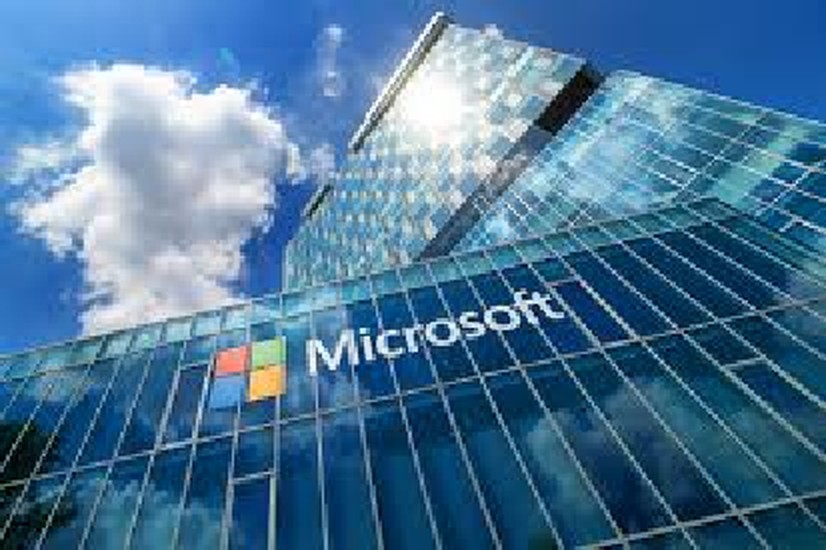}
\end{center}\vspace{-2mm}
\caseq{Question.}\\ \textit{``The image shows a company building. This company officially announced its acquisition of a developer platform in 2018. Is the tail on the platform's logo on the left or right?''}\\
\caseq{Ground truth.} Left.\\[3pt]
\caseq{Static World Growth} \nomark\\
\tool{visual\_search}(image)\\
\textit{Obs:} An office tower of Microsoft.\\
\textit{Think:} Microsoft bought GitHub in 2018. Its logo is a cat, and I recall the tail on the right.\\
\textit{Answer:} Right. \nomark\\[3pt]
\caseq{Search-RSI} \okmark\\
\tool{visual\_search}(image, region = sign on the facade)\\
\textit{Obs:} Microsoft logo and wordmark.\\
\tool{text\_search}(``Microsoft 2018 developer platform'')\\
\textit{Obs:} Microsoft announced its acquisition of GitHub on June 4, 2018.\\
\textit{Think:} The orientation of the tail is a detail I should check on the logo itself.\\
\tool{lookup}(``GitHub'', ``logo'')\\
\textit{Obs:} The GitHub mark, a silhouette of the Octocat whose tail curls out at the lower left.\\
\textit{Answer:} Left. \okmark
\end{small}
\end{tcolorbox}

\textbf{Analysis.} Both agents recover the company and the platform, so the difference lies in the last hop. The baseline answers the visual attribute from memory, the early answering pattern. In the training world of static growth, many final hop facts of this kind are absent from the index, so a lookup that would have found them is often rewarded with nothing, and the habit of checking is not reinforced. Support injection keeps such lookups rewarded for every committed task, and the trajectory of \ours retrieves the logo before answering.

\subsection{Case 2: Grounding the Right Region}
The second case refers to one of two satellites in an illustration, so the agent must search the right region before any text search can help.

\begin{tcolorbox}[title=Case 2: VisBrowse item 129 (Technology),width=\linewidth,breakable,colback=white,colframe=black!55,fonttitle=\small\bfseries,left=3pt,right=3pt]
\begin{small}
\begin{center}
\includegraphics[width=0.86\linewidth]{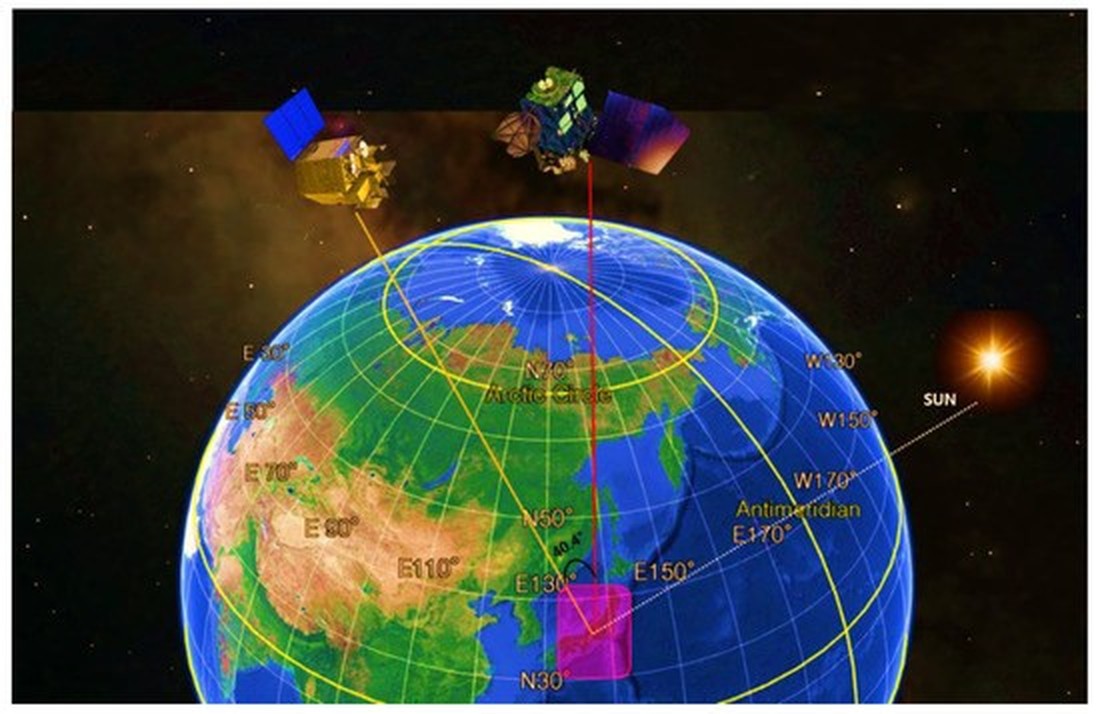}
\end{center}\vspace{-2mm}
\caseq{Question.} \textit{``The satellite on the right side of the image captured an explosion event in 2015. In the Wikipedia ball-and-stick model of the explosive that was most abundant in that explosion, how many ammonium ions are there?''}\\
\caseq{Ground truth.} 12.\\[3pt]
\caseq{Static World Growth} \nomark\\
\tool{visual\_search}(image)\\
\textit{Obs:} Illustrations of satellites observing East Asia.\\
\textit{Think:} Several satellites could have seen an explosion in 2015. Explosions are often caused by ammonium nitrate, whose model shows a few ions.\\
\textit{Answer:} 4. \nomark\\[3pt]
\caseq{Search-RSI} \okmark\\
\textit{Think:} The question refers to the satellite on the right, so I search that region only.\\
\tool{visual\_search}(image, region = right satellite)\\
\textit{Obs:} Himawari-8, the geostationary weather satellite of the Japan Meteorological Agency.\\
\tool{text\_search}(``Himawari-8 explosion 2015'')\\
\textit{Obs:} Himawari-8 imaged the explosions at the Port of Tianjin on August 12, 2015.\\
\tool{lookup}(``2015 Tianjin explosions'', ``hazardous materials'')\\
\textit{Obs:} The warehouse stored about 800 tonnes of ammonium nitrate, the largest quantity among the hazardous materials.\\
\tool{lookup}(``Ammonium nitrate'', ``ball-and-stick model'')\\
\textit{Obs:} The ball-and-stick model of the crystal structure; counting the ammonium ions in the model gives 12.\\
\textit{Answer:} 12. \okmark
\end{small}
\end{tcolorbox}

\textbf{Analysis.} The baseline searches the whole image, never resolves which satellite is meant, and guesses the rest of the chain, which is the coarse grounding pattern followed by early answering. \ours searches the region the question points to and then resolves each of the four hops with a retrieval. Such long chains sit beyond the frontier of the round 0 policy, and they enter training mostly through self demonstration, which forces every hidden entity to appear in an observation before it is used.

\subsection{Case 3: Committing to the Right Entity}
The third case starts from a page of a handwritten score. A plausible but wrong composer leads to a chain that still looks consistent, because the wrong composer is also on the record that the question mentions.

\begin{tcolorbox}[title=Case 3: VisBrowse item 78 (Art),width=\linewidth,breakable,colback=white,colframe=black!55,fonttitle=\small\bfseries,left=3pt,right=3pt]
\begin{small}
\begin{center}
\includegraphics[width=0.78\linewidth]{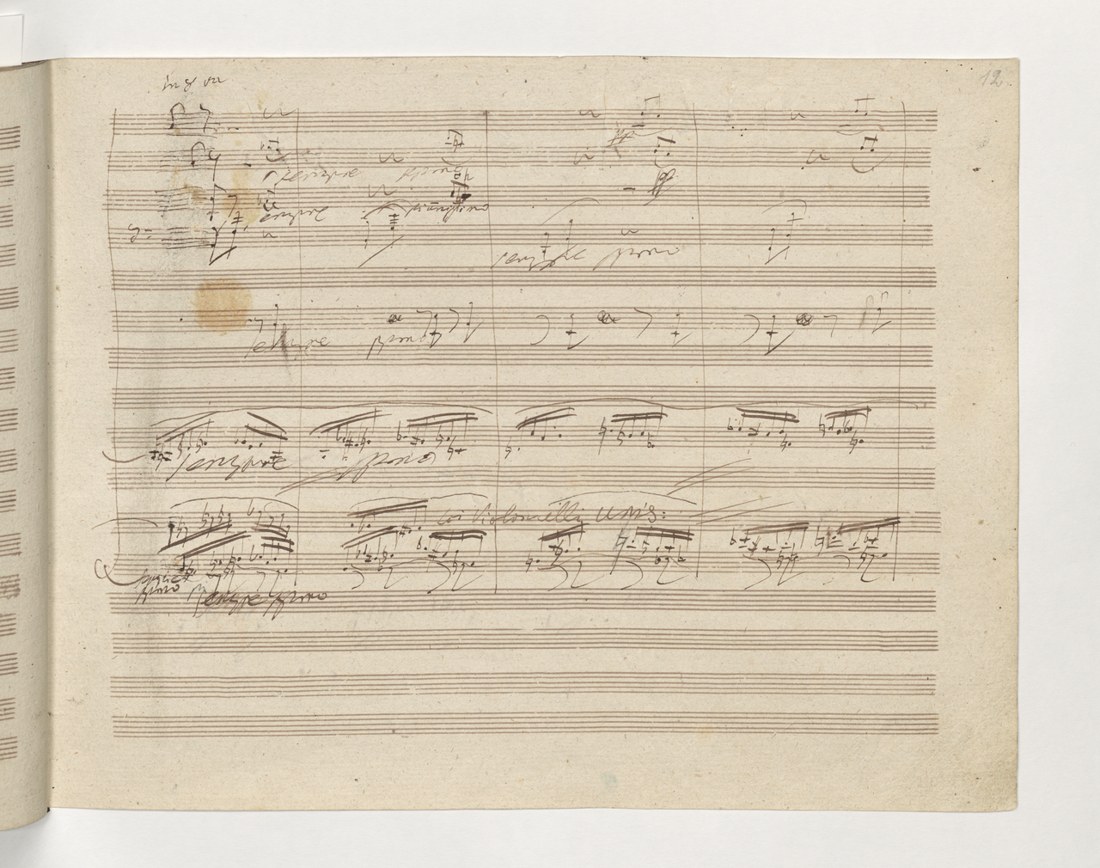}
\end{center}\vspace{-2mm}
\caseq{Question.} \textit{``Who is the author of this musical score manuscript? This composer's work was included on a gold-plated record and sent into space in 1977. How many circles are there on this gold record besides the central hole?''}\\
\caseq{Ground truth.} Beethoven; 7.\\[3pt]
\caseq{Static World Growth} \nomark\\
\textit{Think:} A dense classical manuscript; Mozart's autographs are the most widely reproduced.\\
\textit{Think:} Mozart is on the Voyager Golden Record, launched in 1977, and the record shows a few concentric circles.\\
\textit{Answer:} Mozart; 3. \nomark\\[3pt]
\caseq{Search-RSI} \okmark\\
\tool{visual\_search}(image)\\
\textit{Obs:} Matches with digitized autograph manuscripts of Ludwig van Beethoven.\\
\tool{text\_search}(``Voyager Golden Record Beethoven'')\\
\textit{Obs:} The 1977 Voyager Golden Record includes the first movement of Beethoven's Symphony No.~5 and the Cavatina of his String Quartet No.~13.\\
\tool{lookup}(``Voyager Golden Record'', ``image'')\\
\textit{Obs:} Photograph of the gold plated record; counting the circles other than the central hole gives 7.\\
\textit{Answer:} Beethoven; 7. \okmark
\end{small}
\end{tcolorbox}

\textbf{Analysis.} The baseline never retrieves the composer and commits to a candidate that happens to satisfy the second condition as well, so nothing in its chain signals the error. \ours grounds the manuscript first and then retrieves the record rather than recalling it. Tasks of this kind, where the first hop decides everything, have high hop level disagreement $U(x)$ at round 0, because rollouts agree on the record but disagree on the composer, so admission by $U(x)$ brings them into training with priority.

\subsection{Case 4: Keeping Every Constraint}
The fourth case chains four conditions across two museums and two painters, and a well known painting with a similar title is a tempting distractor.

\begin{tcolorbox}[title=Case 4: VisBrowse item 60 (Art),width=\linewidth,breakable,colback=white,colframe=black!55,fonttitle=\small\bfseries,left=3pt,right=3pt]
\begin{small}
\begin{center}
\includegraphics[width=0.42\linewidth]{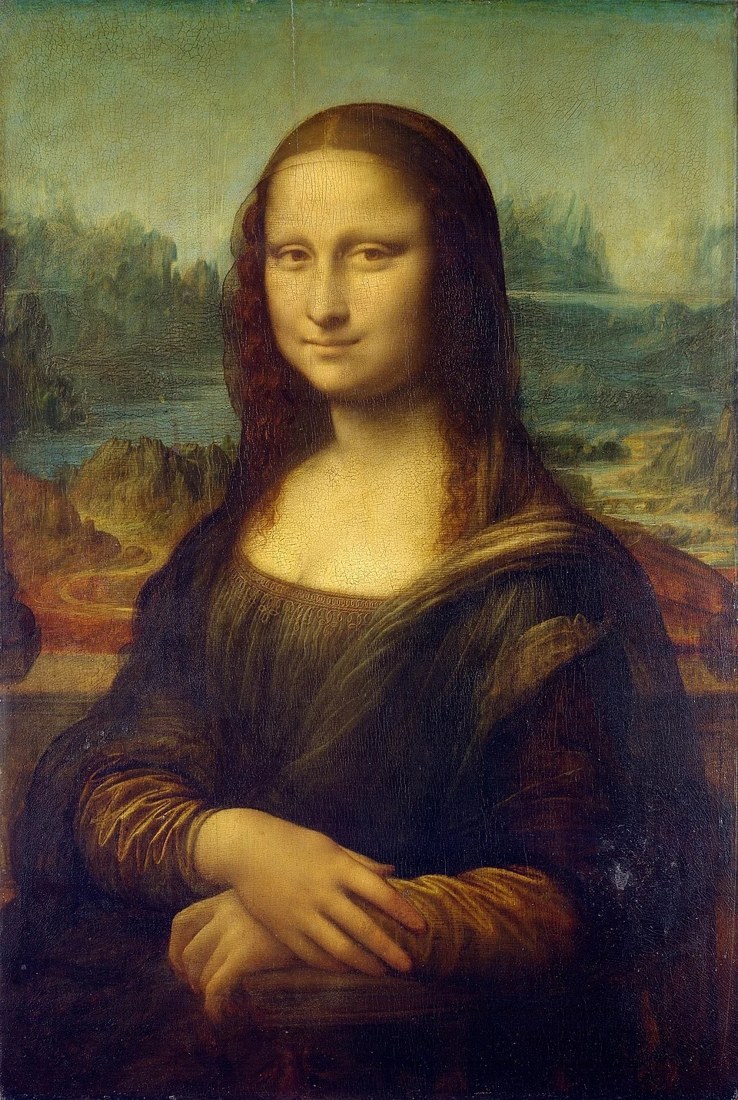}
\end{center}\vspace{-2mm}
\caseq{Question.} \textit{``The place that collects the painting in the picture also has a painting of an emperor's coronation. The painter of this painting also painted a work of this emperor crossing a mountain range. In all versions of this work, how many soldiers are there between the horses' hooves in the foreground?''}\\
\caseq{Ground truth.} 2.\\[3pt]
\caseq{Static World Growth} \nomark\\
\tool{visual\_search}(image)\\
\textit{Obs:} Mona Lisa, Louvre Museum.\\
\textit{Think:} An emperor crossing a mountain range is Napoleon crossing the Alps. In the version where he rides a mule there are no soldiers under the hooves.\\
\textit{Answer:} 0. \nomark\\[3pt]
\caseq{Search-RSI} \okmark\\
\tool{visual\_search}(image)\\
\textit{Obs:} Mona Lisa, Louvre Museum.\\
\textit{Think:} I must first find the coronation painting in the Louvre, because its painter fixes which crossing painting is meant.\\
\tool{text\_search}(``Louvre coronation of emperor painting'')\\
\textit{Obs:} \textit{The Coronation of Napoleon} by Jacques-Louis David, in the Louvre.\\
\tool{text\_search}(``Jacques-Louis David Napoleon crossing the Alps versions'')\\
\textit{Obs:} David painted \textit{Napoleon Crossing the Alps} in five versions, all showing Napoleon on a rearing horse.\\
\tool{lookup}(``Napoleon Crossing the Alps'', ``foreground'')\\
\textit{Obs:} Images of the five versions; counting the soldiers between the hooves in the foreground gives 2 in every version.\\
\textit{Answer:} 2. \okmark
\end{small}
\end{tcolorbox}

\textbf{Analysis.} The baseline drops the second condition, that the crossing painting must be by the painter of the coronation, and answers for the later painting by Paul Delaroche, in which Napoleon rides a mule. This is the constraint dropping pattern. \ours resolves the painter before looking at any crossing painting, the order that anchor first self demonstrations enforce, and it checks the attribute on the versions themselves.